\documentclass[final,5p,times]{elsarticle}
\usepackage{graphicx}
\usepackage{subcaption}
\usepackage{epsfig}
\usepackage{siunitx}
\usepackage{booktabs}
\usepackage[T1]{fontenc}
\usepackage{multirow} 
\usepackage{tabularx}
\usepackage{array}
\usepackage[namelimits]{amsmath} 
\usepackage{amssymb} 
\usepackage{amsmath}            
\usepackage{amsfonts}            
\usepackage{mathrsfs} 
\usepackage{float}
\usepackage{cases}
\usepackage{wrapfig}
\usepackage{hyphenat}
\usepackage{multicol}
\usepackage{caption}
\usepackage[labelfont=bf, textfont=normalfont, justification=raggedright, singlelinecheck=false]{caption}
\usepackage{graphicx}
\usepackage{subcaption}
\usepackage{tabularx}  
\usepackage{amsmath}

\usepackage{amssymb}

\journal{arxiv}
\date{}

\begin{document}
\begin{frontmatter}



\title{ Region Feature Descriptor Adapted to High Affine Transformations \tnoteref{13}}


\tnotetext[13]{This work was supported in part by the National Natural Science Foundation of China (No. 62172190), National Key Research and Development Program(No. 2023YFC3805901), the "Double Creation" Plan of Jiangsu Province (Certificate: JSSCRC2021532) and the "Taihu Talent-Innovative Leading Talent" Plan of Wuxi City(Certificate Date: 202110). }
\author[label1]{Shaojie Zhang}
\ead{7213107006@stu.jiangnan.edu.cn}
\author[label1,label2]{Yinghui Wang\corref{cor1}}
\ead{wangyh@jiangnan.edu.cn}
\author[label1]{Bin Nan}
\ead{binnan98@qq.com}
\author[label1]{Wei Li}
\ead{cs_weili@jiangnan.edu.cn}
\author[label1]{Jinlong Yang}
\ead{yjlgedeng@163.com}
\author[label1]{Tao Yan}
\ead{yantao.ustc@gmail.com}
\author[label1]{Yukai Wang}
\ead{ericwangyk22@163.com}
\author[label3]{Liangyi Huang}
\ead{lhuan139@asu.edu}
\author[label4]{Mingfeng Wang}
\ead{mingfeng.wang@brunel.ac.uk}
\author[label5]{Ibragim R. Atadjanov}
\ead{ibragim.atadjanov@gmail.com}
\cortext[cor1]{Corresponding author}
\affiliation[label1]{organization={ School of Artificial Intelligence and Computer Science, Jiangnan University},
            addressline={1800 Li Lake Avenue},
            city={wuxi},
            postcode={214122},
            state={Jiangsu},
            country={PR China}}
\affiliation[label2]{organization={ Engineering Research Center of Intelligent Technology for Healthcare, Ministry of Education},
            addressline={1800 Li Lake Avenue},
            city={wuxi},
            postcode={214122},
            state={Jiangsu},
            country={PR China}}
 \affiliation[label3]{organization={School of Computing and Augmented Intelligence, Arizona State University},
            addressline={1151 S Forest Ave},
            city={Tempe},
            postcode={8528},
            state={AZ},
            country={U.S}}
\affiliation[label4]{organization={Department of Mechanical and Aerospace Engineering, Brunel University},
            addressline={Kingston Lane},
            city={London},
            postcode={UB8 3PH},
            state={Middlesex},
            country={U.K}}    
\affiliation[label5]{organization={Tashkent University of Information Technologies named after al-Khwarizmi},
			 addressline={ 108 Amir Temur Avenue},
		     city={Tashkent},
			postcode={100084},
			 state={},
			 country={Uzbekistan}}

\begin{abstract}
To address the issue of feature descriptors being ineffective in representing grayscale feature information when images undergo high affine transformations, leading to a rapid decline in feature matching accuracy, this paper proposes a region feature descriptor based on simulating affine transformations using classification. The proposed method initially categorizes images with different affine degrees to simulate affine transformations and generate a new set of images. Subsequently, it calculates neighborhood information for feature points on this new image set. Finally, the descriptor is generated by combining the grayscale histogram of the maximum stable extremal region to which the feature point belongs and the normalized position relative to the grayscale centroid of the feature point's region. Experimental results, comparing feature matching metrics under affine transformation scenarios, demonstrate that the proposed descriptor exhibits higher precision and robustness compared to existing classical descriptors. Additionally, it shows robustness when integrated with other descriptors.
\end{abstract}



\end{frontmatter}


\section{Introduction}
Feature descriptors are crucial for distinguishing differences between different feature points in the same image and preserving identical information for the same feature points in different images. However, under affine transformations, especially when the tilt angle between the optical axis and the scene is greater than 40 degrees, the neighborhood grayscale information used to describe feature points undergoes significant changes. This leads to difficulties in maintaining the distinctiveness and invariance of descriptors.

The challenges stem from the deformation caused by affine transformations, where the region information corresponding to the descriptor of the same feature point undergoes substantial changes, making it challenging to maintain invariance. Additionally, existing descriptors often represent the smaller neighborhood around feature points, and the limited information in these small regions may result in difficulties in maintaining the distinctiveness of descriptors under affine transformations. Common methods such as SIFT (Scale Invariant Feature Transform) \cite{ref1}, BRIEF (Binary Robust Independent Elementary Features) \cite{ref2}, and AKAZE (Accelerated KAZE) \cite{ref3} struggle to achieve the desired descriptor performance under affine transformations.

To address the invariance issue, current strategies \cite{ref4}-\cite{ref6} involve simulating image transformations with different affine degrees to maintain descriptor invariance. However, existing simulation strategies use fixed data for simulating transformations. For instance, the well-known ASIFT (Affine Scale Invariant Feature Transform) algorithm \cite{ref4} simulates only the affine enlargement transformation on images, leading to a larger affine gap between the two sets of affine-transformed image collections. This makes it challenging to ensure the invariance of the same descriptor. Simulating affine transformations separately for matched images with varying degrees of affinities, i.e., decreasing the affinity for highly affine images and increasing it for less affine images, can significantly improve this drawback.

Furthermore, to address the distinctiveness of descriptors, current methods \cite{ref7}-\cite{ref9} use neighborhood information supplementation, but the limited neighborhood information often results in an inability to distinguish the distinctiveness of descriptors. Directly expanding the region of description, while effectively mitigating the similarity problem of descriptors in different large regions, may cause overlapping of region information for descriptors corresponding to nearby regions within the same large region. This seriously affects the distinctiveness of descriptors within the same large region. Observing and judging whether two feature points in two images are the same using the human eye often involves first determining the image region to which the point belongs, then locating the specific position within that region. Finally, a careful examination of the local region features is performed, considering that similar descriptors may be deemed different if differences are found. Utilizing this principle to characterize descriptors can alleviate the problem of the lack of distinctiveness in descriptors under affine scenes caused by the limited neighborhood information.

In summary, this paper addresses the effectiveness drop of common descriptors under significant affine transformations, specifically when simulating a tilt angle of up to 40 degrees between the optical axis and the scene. The proposed solutions focus on maintaining the invariance of descriptors by classifying images with different affine degrees and implementing corresponding shrinking or enlarging affine transformations. This helps reduce the changes in the region representation of the same feature point descriptor. Additionally, the distinctiveness of descriptors is maintained by adding grayscale features of larger regions to the original descriptors, along with the relative position of the feature point to the region centroid. Appropriate weights are assigned to the newly added region information and the original descriptor. The main contributions and advantages of this paper can be summarized as follows:

(1) Proposed a strategy for classifying simulated affine transformations, capable of maintaining the invariance of the same descriptor under significant affine transformations.

(2) Designed a descriptor that adds more region information to the original descriptor, enhancing the distinctiveness of different descriptors when discerning similar small regions undergoing affine changes.

(3) Implemented an evaluation method for descriptor accuracy based on the homography matrix and fundamental matrix of the proposed feature descriptor. This verified the effectiveness and generality of the proposed method.

\begin{figure*}[t]
    \centering
    \includegraphics[height=3in]{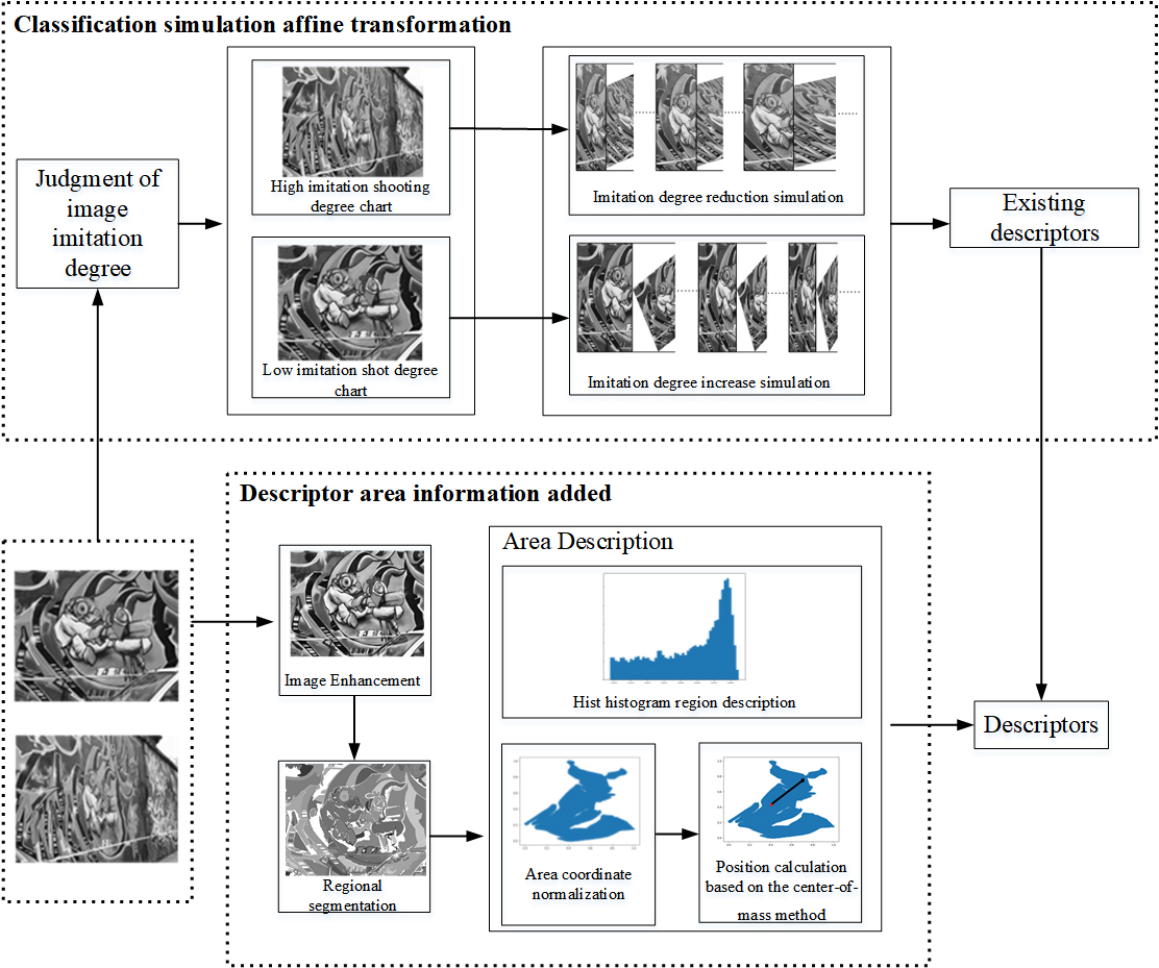}
\caption{Methodological Framework.}
    \label{Fig1}
\end{figure*}

\section{RELATED WORK}
Descriptors are used to represent the grayscale information of the neighborhood of a feature point, calculating pixel intensity or intensity variations within the neighborhood and mapping them into a vector of floating-point values or binary values. The research on feature descriptors has spanned several decades, evolving from early methods like Gaussian derivatives \cite{ref10}, moments invariants \cite{ref11}, and affine texture invariants \cite{ref12} to the more mature descriptor systems in use today. However, these descriptors still face challenges in maintaining invariance and distinctiveness under affine transformations, particularly in the case of significant affine changes. Generally, common descriptors under affine transformations fall into two categories: those based on gradient statistics and those based on local intensity comparisons.

Among descriptors based on gradient statistics, the most classical method is the descriptor constructed in Lowe's SIFT (Scale Invariant Feature Transform) \cite{ref1}. This method divides the descriptor's representation of the neighborhood, under normalized scale and rotation, into a $4\times4$ grid. It computes the gradient direction histogram for each grid, enhancing the robustness of the descriptor to image rotation and scale changes, making it adaptable to smaller affine transformations. Other methods based on gradient statistics have been developed by improving upon the SIFT descriptor. For instance, to speed up computation, Bay et al. proposed the SURF (Speeded Up Robust Features) method \cite{ref13}, which simplifies the SIFT descriptor and calculates horizontal and vertical Haar wavelet responses within the neighborhood to enhance the effectiveness of the descriptor. To adapt the descriptor's neighborhood to affine transformations, Tola et al. introduced the DAISY descriptor \cite{ref7}, changing the rectangular neighborhood of the SIFT-based descriptor to a circular, centrally symmetric structure. However, even with circular neighborhoods, the images under affine transformations may become ellipses, and circular neighborhoods still cannot fully adapt to affine transformations. The DSP-SIFT method \cite{ref8}, proposed by Dong et al., extended the descriptor's neighborhood to multiple scales in the scale pyramid, collecting gradient statistics for the same neighborhood across multiple scales. Similarly, to maintain the invariance of the descriptor's neighborhood under affine transformations, Morel et al. introduced the ASIFT algorithm \cite{ref4}, which simulates affine transformations on images, ensuring the invariance of the grayscale information within the descriptor's neighborhood. While ASIFT significantly improves descriptor effectiveness under affine transformations, the simulation is limited, and it does not fully address the issues in affine scenes. To address the problem of parameter setting in simulated image models, Cai et al. proposed a perspective transformation simulation based on a particle swarm optimization algorithm \cite{ref5}, and Zeng et al. introduced the IABC-PSIFT algorithm \cite{ref6}, which combines a perspective transformation model with an artificial bee colony algorithm to obtain optimal simulation parameters. However, the effectiveness of these methods is still limited.

Descriptors based on local intensity comparisons are exemplified by the BRIEF (Binary Robust Independent Elementary Features) descriptor introduced by Calonder et al. \cite{ref2}. In this method, a fixed number of sampled points is chosen within the descriptor's corresponding neighborhood based on the Cartesian coordinates following an isotropic Gaussian distribution. The binary encoding of the intensity values of the points is used as the descriptor. To adapt to affine scenes, improvements in the strategies for selecting sampling points are the primary focus among descriptors based on local intensity comparisons. The ORB (Oriented FAST and Rotated BRIEF) method \cite{ref14}, proposed by Rublee et al., constructs the descriptor based on the BRIEF descriptor. It incorporates a scale pyramid and the grayscale centroid method and uses machine learning to select robust sampling point pairs, enhancing the robustness of the descriptor to scale and rotation changes. Subsequently, Leutenegger et al. introduced the BRISK (Binary Robust Invariant Scalable Keypoints) method \cite{ref15}, which constructs the descriptor by sampling points on concentric circles with varying radii. Alahi et al. proposed the FREAK (Fast Retina Keypoint) descriptor \cite{ref9}, inspired by the characteristics of the retina receiving images, which also employs a strategy of selecting sampling points on concentric circles with varying radii. The main difference is that the BRISK method constructs the descriptor by selecting a fixed number of sampling points on circles with different radii, while the FREAK descriptor, to increase computational speed, focuses on information near the neighborhood center, reducing the number of sampling points as the radius increases. This makes the descriptor constructed by the BRISK method contain more grayscale information than the FREAK descriptor, and under affine transformations, it is easier for the BRISK descriptor to select matching point pairs for comparison than the FREAK descriptor. However, both methods still face challenges when most sampling points are offset due to significant affine transformations, affecting the preservation of descriptor characteristics under such changes.

In summary, existing methods construct descriptors based on the neighborhood of feature points. The inability of descriptors to maintain invariance under affine scenes is primarily attributed to the inadequacy of neighborhood selection strategies in adapting to affine transformations. Therefore, this paper proposes a descriptor construction method that is more suitable for affine scenes. It involves the classification and simulation of affine transformations, either shrinking or enlarging, for images with relatively high or low affine degrees. Subsequently, existing descriptor construction methods are used to compute and add the grayscale histogram probability of the larger region to which the feature point belongs, as well as the position information of the feature point relative to the centroid of the region.

\section{METHODOLOGY}
To address the challenges posed by significant affine transformations with tilt angles up to 40 degrees, causing variations in the region information and hindering the maintenance of descriptor consistency and distinctiveness, this paper proposes a method of simulating affine transformations for constructing a region information descriptor. The technical pipeline of this method is illustrated in Figure 1 and consists of two main components: classification-simulated affine transformations and the addition of descriptor region information.

In the classification-simulated affine transformations, the degree of affine transformation for the original image is assessed. Based on this assessment, a relative high/low degree of affine transformation is determined, and simulated transformations are implemented accordingly. Feature point descriptors are then computed on the obtained set of simulated images.

Simultaneously, in the addition of descriptor region information, the input images to be matched are preprocessed for image enhancement. Region segmentation is performed using the Maximally Stable Extremal Regions (MSER) algorithm \cite{ref16}, and the region is characterized using a grayscale histogram. Subsequently, based on normalized region coordinates, the gray\\scale centroid method is employed to rotate the main direction of the region to align with the positive x-axis. The relative position of the feature point to the centroid after normalization is also calculated.

Finally, corresponding weights are assigned to the region description information and relative position information. These weighted components are then fused with the original feature point descriptor to obtain the feature descriptor proposed in this paper.

\subsection{Classification-Simulated Affine Transformation}
The fundamental model for the classification-simulated affine transformation in this paper is based on the ASIFT method \cite{ref4}. This includes the basic decomposition of the affine matrix and mapping the decomposed parameters to camera three-dimensional pose transformations. An affine matrix represents a composite of non-singular linear transformation and translation, as shown in Equation (1). It comprises the linear component A of the affine matrix and the translation vector b. The linear component A involves a combination of rotation and non-uniform scaling, where non-uniform scaling is the key distinction between affine and isometric transformations, as illustrated in Equation (2).

\begin{equation}
	\label{eq1}
	x=H_{A}x=\begin{bmatrix}
		A & b\\ 
		0^{T} & 1
		\end{bmatrix}x
\end{equation}
 
\begin{multline}
	\label{eq2}
	A=\lambda R(\Psi )D(t)R(\phi )=\\\lambda \begin{bmatrix}
		cos\Psi  & -sin\Psi  \\ 
		 sin\Psi & cos\Psi 
		\end{bmatrix}\begin{bmatrix}
		t & 0 \\ 
		0 & 1
		\end{bmatrix}\begin{bmatrix}
		cos\phi  & -sin\phi \\ 
		 sin\phi& cos\phi
		\end{bmatrix}
\end{multline}
Here, each parameter can be found with corresponding meanings in Figure 2. The plane $u$ represents the plane where the scene is captured, and point $O$ is the intersection of the camera optical axis with plane $u$. $\lambda $denotes the image scaling factor. $\Psi$  is the angle of rotation around the optical axis of the camera, determining the rotation angle in the affine transformation of the 2D image. $\Phi$ represents the angle between the projection line of the optical axis onto plane u and the positive x-axis direction. The angle $\Phi$ is positive in the positive y-axis direction and negative in the negative direction, determining the direction of non-uniform scaling in the affine transformation. t is the affine quantity, determining the magnitude of non-uniform scaling in the affine transformation. It holds that $t=1/cos\theta$, where $\theta$  is the angle between the optical axis and the normal vector to plane $u$, representing the tilt angle of the optical axis.
\begin{figure}[!hptb]
    \begin{center}
    \includegraphics[width=2.1in]{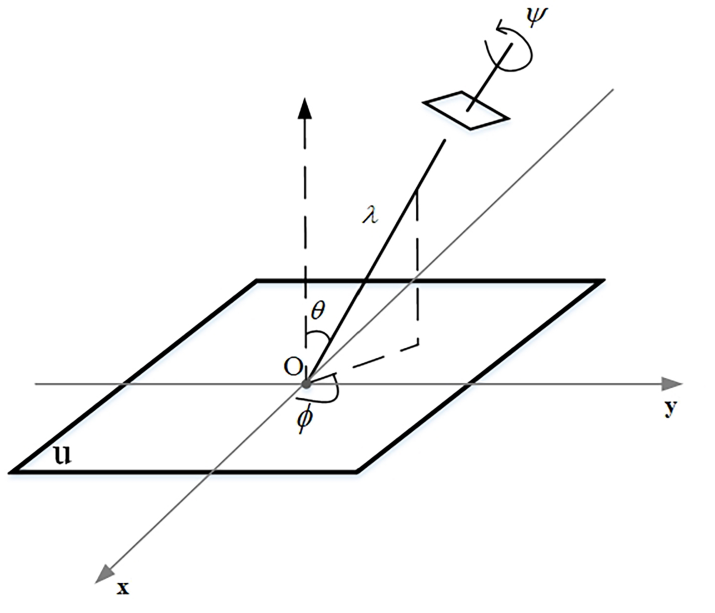}
    \end{center}
\caption{Affine Transformation Parameters Interpretation Chart.} \label{Fig2}
\end{figure}

Based on the above decomposition, simulating affine transformations in images can be achieved by setting parameters $\lambda $, angles $\Psi$, $\Phi$, and $\theta$. Among these parameters, $\lambda$and $\Psi$ influence the scale and rotation transformations in the image. Both of these parameters are minimized in specific descriptors through scale-space simulation and rotation-invariant corresponding operations. Therefore, there is no need to simulate them explicitly in this context. Only the core parameters in affine transformations need to be simulated, i.e., angles $\phi$ and $\theta$, which influence the non-uniform scaling direction and size in affine transformations. The larger the angle $\Phi$, the greater the corresponding affine quantity t, resulting in a more pronounced affine effect in the final image. When the initial camera has a certain angle $\theta$, reducing $\theta$ can decrease the degree of affine transformation in the original image. Correspondingly, only the parameter t in Equation (2) needs to be set to an appropriate value less than 1.

In the ASIFT algorithm, the default input image pair corresponds to $\theta$ angle of $\theta$ degrees. Therefore, for simulated images undergoing simulated transformations, $\theta$ can only be increased, leading to an increased degree of affine transformation in the image. The key difference in this paper's approach is that it assumes the input image corresponds to a certain initial $\theta$ angle. Consequently, the degree of affine transformation in the image can be reduced by decreasing $\theta$ for highly affine images, and conversely, increasing $\theta$ can enhance the degree of affine transformation. This approach aims to generate a set of simulated images that are closer to the intended goal.

To simulate affine transformations for classification, it is necessary to first determine the degree of affine transformation between images in a given image pair. Experiments have shown that reducing the difference in the degree of affine transformation between images is beneficial for increasing the number of feature matches, while the opposite may result in a decreased number of feature matches. Therefore, by performing cross-feature matching between the two input images and simulated images with increased affine transformation degrees, the relative degree of affine transformation between images can be assessed by comparing the number of feature matches. Specifically, for input images a and b, simulate the addition of an affine quantity t, generating images A and B. Using the SIFT feature extraction and description algorithm for matching, if the number of feature matches between images a and B is denoted as  $M_{aB}$, and the number of feature matches between images A and b is denoted as $M_{Ab}$, then if $M_{aB}$>$M_{Ab}$, image a has a lower degree of affine transformation. Conversely, if $M_{aB}$>$M_{aB}$, image b has a lower degree of affine transformation.

The simulation strategy in this paper is as follows: for images with lower degrees of affine transformation in the input images, a simulated set of enlarging affine quantities is used, $\left \{ \sqrt{2},2,2\sqrt{2}\right \}$; for images with higher degrees of affine transformation in the input images, a simulated set of reducing affine quantities is used, $\left \{ \sqrt{2}/2,1/2,\sqrt{2}/4\right \}$. Both sets are simulated for $\phi$ angles ranging from -45 degrees to 45 degrees with a step size of 15 degrees. In contrast, the ASIFT algorithm uniformly simulates both the affine quantity t and $\phi$ angle, where the sampling set for t is $\left \{ \sqrt{2},2,2\sqrt{2},4,4\sqrt{2}\right \}$, and the simulation for $\phi$ angle ranges from 0 degrees to 180 degrees with a step size of $72/t$ degrees.

To illustrate the advantages of the simulation strategy in this paper, Equations (3) and (4) are introduced.

\begin{equation}
	\label{eq3}
	max\_affine=\frac{max(t_{2})}{min(t_{1})}
\end{equation}
Where $max\_affine\_affine$ represents the maximum simulated affine quantity, and the max() and min() functions correspond to the maximum and minimum affine quantities in the sampling set, respectively. 

\begin{equation} 
	\label{eq4}
	average\_ dffer=\frac{\sum t_{2}a}{count(t_{2})}-\frac{\sum t_{1}}{count(t_{1})}
\end{equation}	
Where $average\_differ$ is the overall average affine quantity, a represents that the affine quantity for high-affine images is a times that of low-affine images, and the count() function calculates the number of elements in the sampling set.

Through Equation (3), it can be observed that the maximum simulated affine quantity in this paper can reach 8, while the ASIFT algorithm only has $4\sqrt{2}$. This indicates that the simulated images in this paper exhibit a higher degree of affine transformation than the ASIFT algorithm, making them suitable for scenes with larger affine transformations. Plugging the sampling set into Equation (4) allows us to calculate the overall average affine quantity difference for the simulated affine images. Through computation, it is evident that the overall average affine quantity difference in this paper is significantly lower than that of the ASIFT algorithm. This implies that the simulated generated affine images are more similar, making it more conducive to maintaining smaller changes in descriptors' corresponding neighborhoods during affine transformations.

The $\phi$ angle determines the direction of non-uniform scaling applied to the image. As per Equation (2), when the $\phi$ angle on the xy coordinate axes in Figure 2 satisfies the diagonal relationship, the direction of non-uniform scaling applied to the image is consistent. The simulation of the $\phi$ angle needs to achieve the following objectives: when the angular difference between two matched images is $\Delta \phi \epsilon [-90^{\circ},90^{\circ}]$ , simulating the $\phi$ angle for both images can generate images in the resulting set with the same $\phi$ angle. Assuming the simulated $\phi$ angle interval for two images is $[i,j]$, it is only necessary to ensure that $[i+\Delta \phi ,j+\Delta \phi ]$ and $[i,j]$ always have an intersection, meaning $j-i\geq90$ to achieve the above objectives. At this point, the simulated interval for the $\phi$ angle is  $[-45^{\circ},45^{\circ}]$. By calculating the total number of simulations in the simulated sampling sets for both this paper's method and the ASIFT algorithm, it is found that this paper simulates only 18 times for a single image, while ASIFT requires 26 simulations. This paper's method can reduce the runtime of simulating affine transformations.

\subsection{Descriptor Region Information Additio}
In this approach, the descriptor is primarily augmented with relevant information about the image region where the feature point is located. Leveraging the invariant properties of pixel grayscale values and the length ratio of parallel line segments in image affine transformations, the MSER algorithm \cite{ref16} is employed for region segmentation in the enhanced image. The grayscale histogram is utilized to characterize the region, and after normalizing the region coordinates, the region is rotated to its main direction using the grayscale centroid method. The position of the feature point relative to the centroid is determined, and finally, both the histogram and relative position are assigned corresponding weights.

\subsubsection{Image Enhancement}
To facilitate region segmentation, it is essential to enhance the contrast and blur details within the same region. Therefore, this paper employs Contrast Limited Adaptive Histogram Equalization (CLAHE) \cite{ref17} and bilateral filtering \cite{ref18} for image enhancement. 

Contrast Limited Adaptive Histogram Equalization (CLAH\\E) divides the image into several sub-block regions. After applying histogram equalization to an individual region using Equation (5), pixels in the histogram of a specific sub-region that exceed the limit are truncated and evenly distributed among the pixel values corresponding to the grayscale levels. The new histogram can be obtained using Equation (6). Here, the new grayscale value is determined by the updated histogram obtained from Equation (5).

\begin{equation} 
	\label{eq5}
	s=255\sum_{x=0}^{r}P(x)=255\sum_{x=0}^{r}\frac{C_{x}}{MN}
\end{equation}	
Where $r$ is the input grayscale value, $s$ is the output grayscale value, $C_{x}$ is the number of pixels with grayscale value $x$, $p(x)$ is the probability distribution of grayscale value $x$, and $M$ and $N$ are the length and width of the subdivided blocks, respectively.

\begin{equation} 
	\label{eq6}
	new\_ c_{x}=\begin{Bmatrix}
		clip & c_{x}+bin\_incr\geq clip \\ 
		c_{x}+bin\_incr & c_{x}+bin\_incr< clip
		\end{Bmatrix}
\end{equation}	
Where $new\_c\_x$ represents the number of pixels corresponding to the re-allocated grayscale level $x$, clip is the clipping threshold, and the increase in pixels for each grayscale level is given by $bin\_incr=clip\_sum/256$, where $clip\_sum$ is the total number of clipped pixels.

In this case, where the differences between the sub-blocks are pronounced, it is necessary to smooth them using bilinear interpolation. The image is divided into the same sub-blocks as described above, with each sub-block further divided into $4\times4$ small regions, as illustrated in Figure 3. By applying equation (7) and referring to the schematic diagram of linear interpolation in Figure 3, the pixel values within the corresponding region can be determined.
\begin{multline} 
	\label{eq7}
	f(x,y)=\frac{y_{2}-y}{y_{2}-y_{1}}[\frac{x_{2}-x}{x_{2}-x_{1}}T_{A}+\frac{x-x_{1}}{x_{2}-x_{1}}T_{B}]+\\\frac{y-y_{1}}{y_{2}-y_{1}}[\frac{x_{2}-x}{x_{2}-x_{1}}T_{C}+\frac{x-x_{1}}{x_{2}-x_{1}}T_{D}]
\end{multline}	
Where the grayscale value $f(x,y)$ represents the sought pixel grayscale value. The grayscale values of the black pixels at the centers of the subdivided regions in Figure 3 are known and denoted as $A(x1,y1)$,$B(x2,y2)$,$C(x3,y3)$,and $D(x4,y4)$ with grayscale values $T_{A}$,$T_{B}$,$T_{C}$,and $T_{D}$, respectively.

\begin{figure}[!hptb]
    \begin{center}
    \includegraphics[width=2.1in]{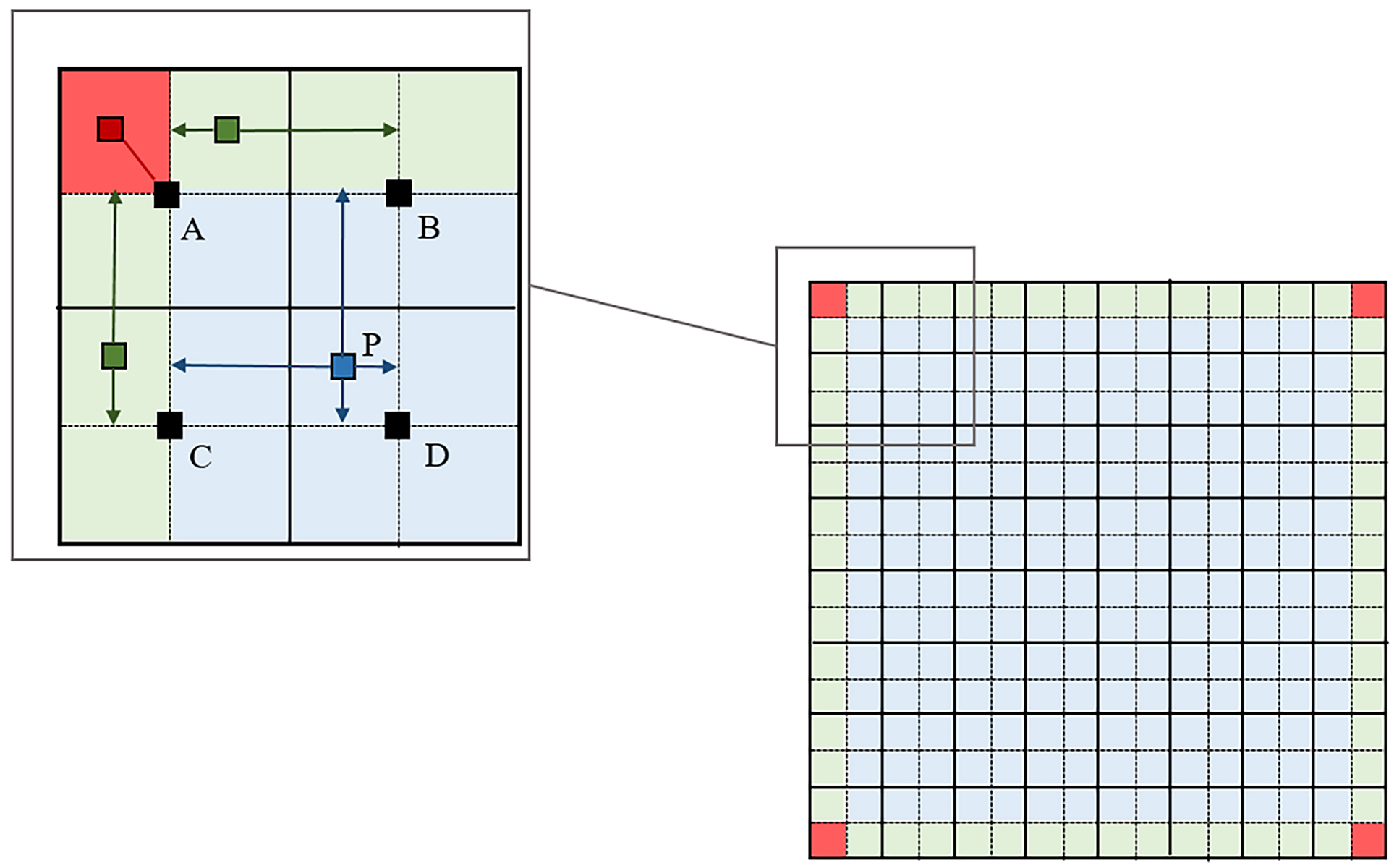}
    \end{center}
\caption{Sub-block segmentation and linear interpolation diagram.} \label{Fig3}
\end{figure}
After applying the Contrast Limited Adaptive Histogram Equalization (CLAHE) algorithm for contrast enhancement in the image, a bilateral filter \cite{ref18} is used to smooth the details in the image regions. The bilateral filter, as shown in equation (8), builds upon Gaussian filtering by incorporating weights based on pixel grayscale values. This approach reduces regional details while preserving edge information, enhancing the overall result.
\begin{align} 
	\label{eq8}
& h(x)= k^{-1}(x)\sum_{ \xi\epsilon S}^{ }f(\xi )c(\xi ,x)s(f(\xi ),f(x))\notag \\
& k(x)=\sum_{ \xi\epsilon S}^{ }f(\xi )c(\xi ,x)s(f(\xi ),f(x))\notag \\
& c(\xi ,x)=e^{-\frac{(\xi -x)^{2}}{2\delta _{d}^{2}}}\notag \\
& s(f(\xi ),f(x))=e^{-\frac{f(\xi)-f(x)^{2}}{2\delta _{r}^{2}}}
\end{align}	

Where $h(x)$is the output grayscale value, $k(x)$ is the normalized weight, $x$ is the input pixel, $S$ is the filtering window corresponding to pixel $x$, $f(x)$ is the grayscale value of pixel $x$, $c(\xi ,x)$ represents the spatial weight, and $s(\xi ,x)$ and $f(x)$ are the grayscale weights. $\delta _{d}$  and  $\delta _{r}$ denote the standard deviations of spatial proximity and grayscale similarity, respectively.

The image enhancement effect is illustrated in Figure 4, where the differences in various regions are more pronounced in the CLAHE image. After applying the bilateral filter, the image becomes smoother in each region, indicating a more uniform distribution of grayscale values within the regions. Figure 5 demonstrates the effect of region segmentation before and after image enhancement. Since the original image is grayscale and many regions have similar grayscale values, subsequent region segmentation algorithms may exclude unstable regions. After image enhancement, there is a clearer contrast between regions, and the regions appear smoother internally. This results in more complete and accurate region segmentation.
\begin{figure}[H]	
	\centering
	\begin{subfigure}{0.42\linewidth}
		\centering
		\includegraphics[width=0.99\linewidth]{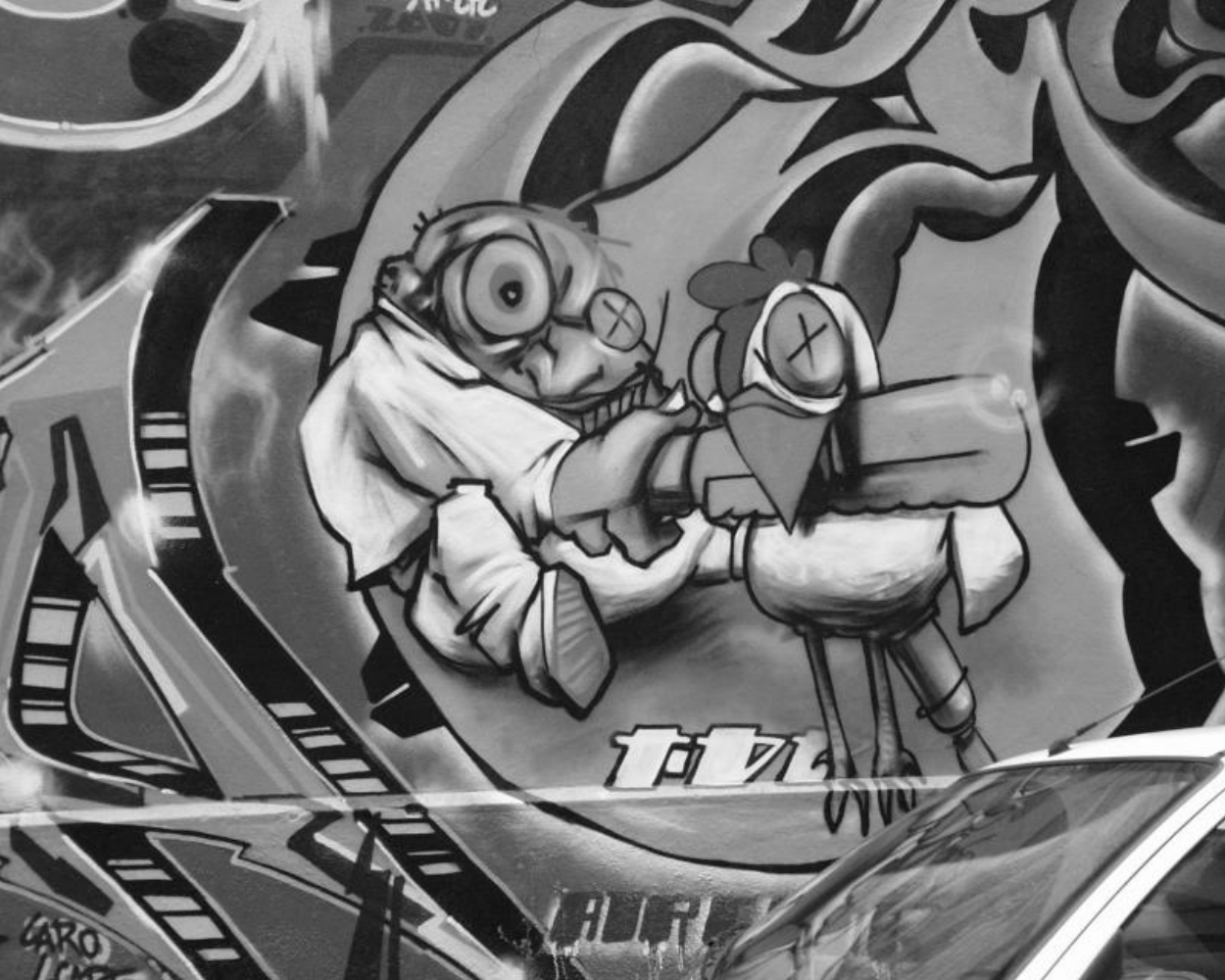}
		\captionsetup{width=.8\linewidth}
		\caption{Original Image}
		\label{Fig41}%
	\end{subfigure}
	\centering
	\begin{subfigure}{0.42\linewidth}
		\centering
		\includegraphics[width=0.99\linewidth]{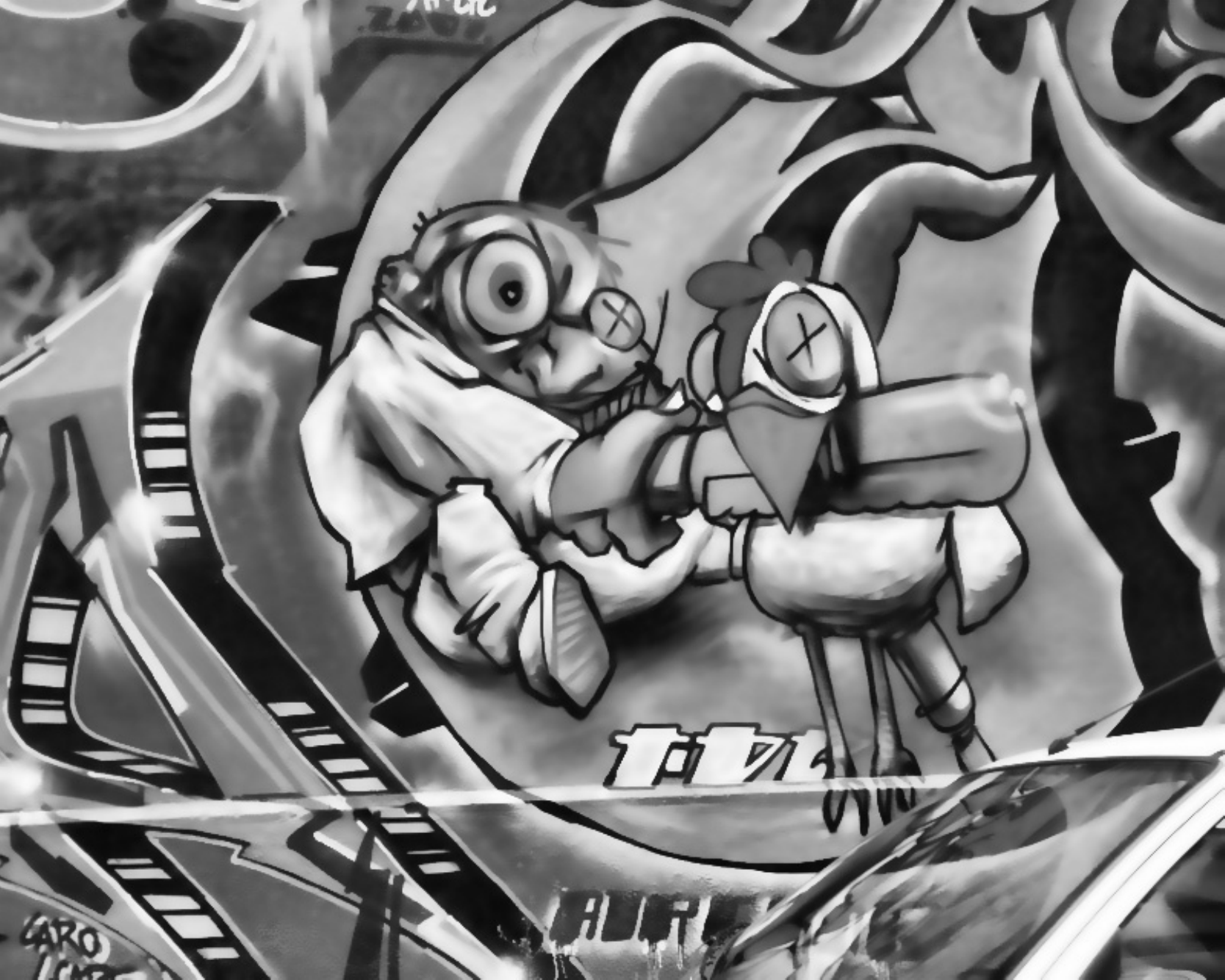}
		\captionsetup{width=.8\linewidth}
		\caption{CLAHE Image   }
		\label{Fig42}
	\end{subfigure}
	\begin{subfigure}{0.42\linewidth}
		\centering
		\includegraphics[width=0.99\linewidth]{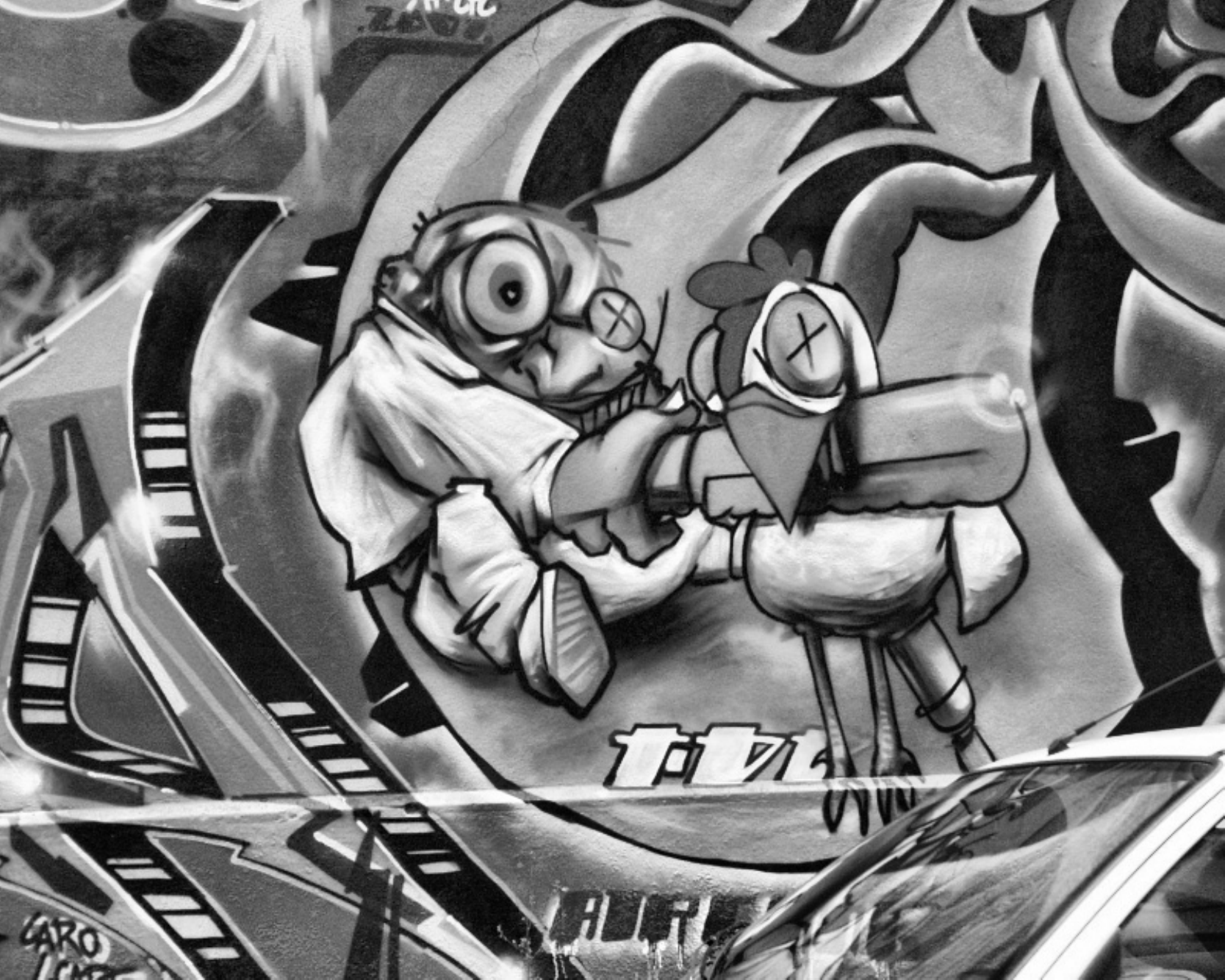}
		\captionsetup{width=.8\linewidth}
		\caption{CLAHE + Bilateral Filter Image}
		\label{Fig43}
	\end{subfigure}
	\caption{Illustration of Image Enhancement Effects.}
\end{figure}

\begin{figure}[H]	
	\centering
	\begin{subfigure}[t]{0.42\linewidth}
		\centering
		\includegraphics[width=0.99\linewidth]{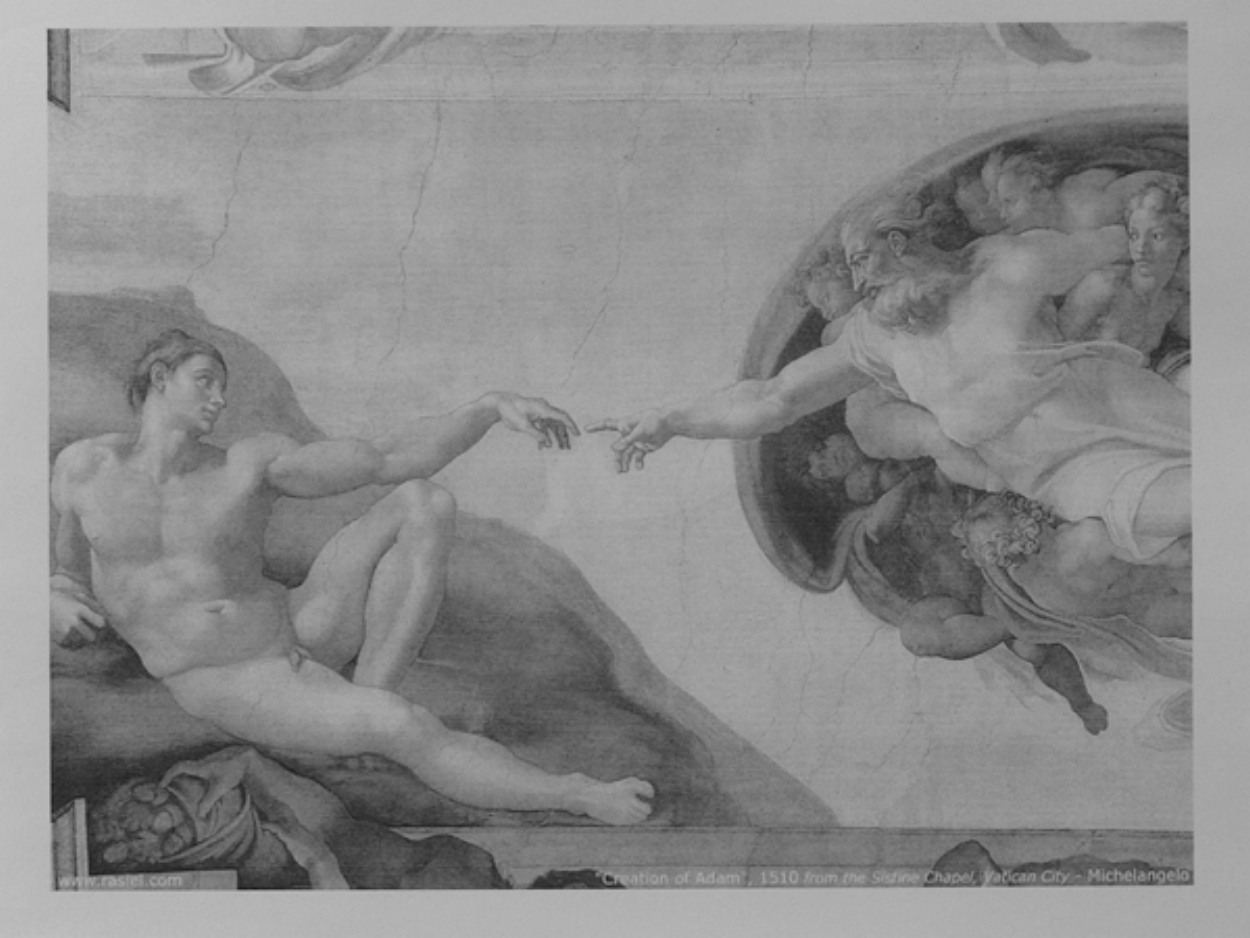}
		\caption{Original Image}
		\label{Fig51}%
	\end{subfigure}
	\centering
	\begin{subfigure}[t]{0.42\linewidth}
		\centering
		\includegraphics[width=0.99\linewidth]{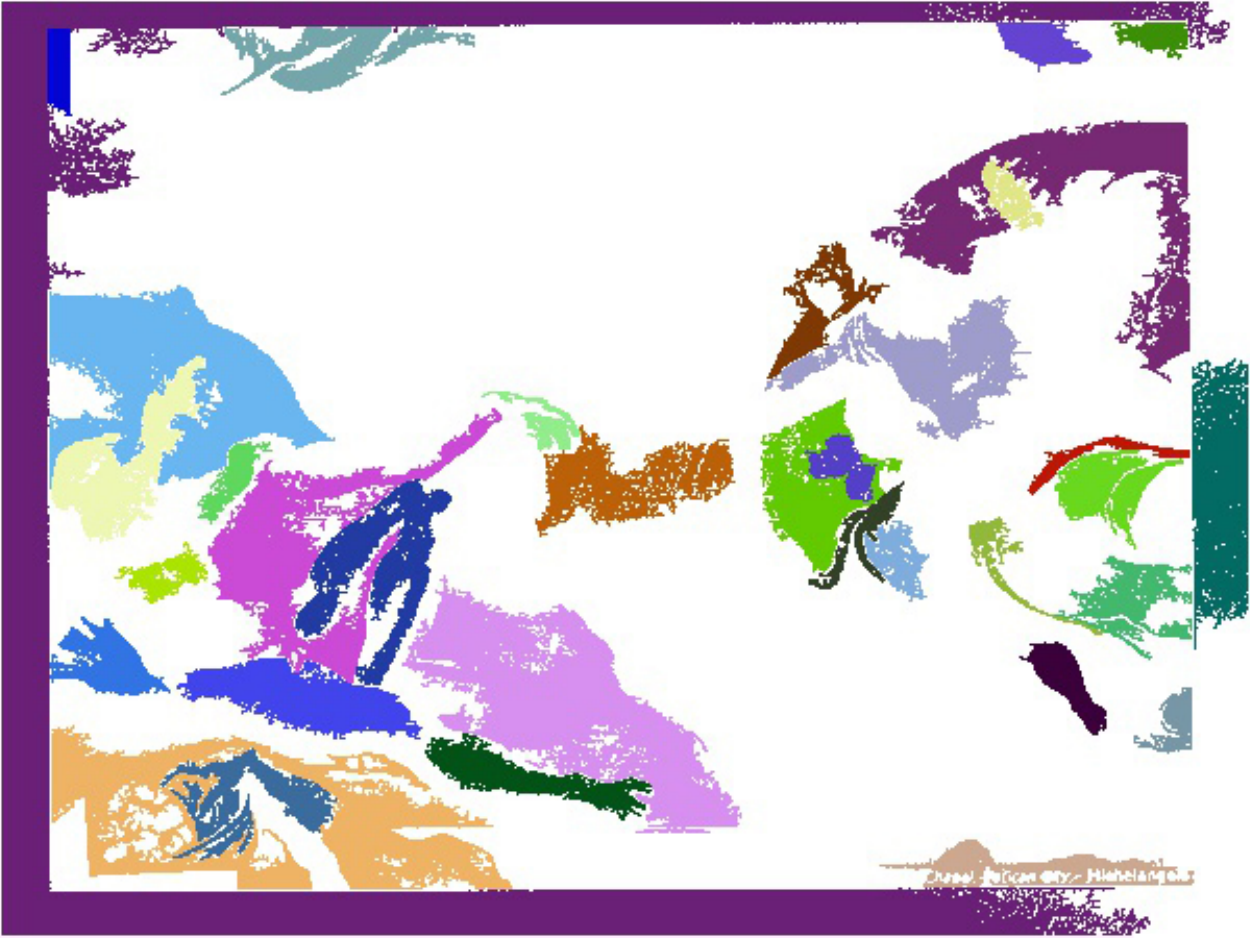}
		\caption{Region Segmentation without Image Enhancement}
		\label{Fig52}
	\end{subfigure}
	\begin{subfigure}[t]{0.42\linewidth}
		\centering
		\includegraphics[width=0.99\linewidth]{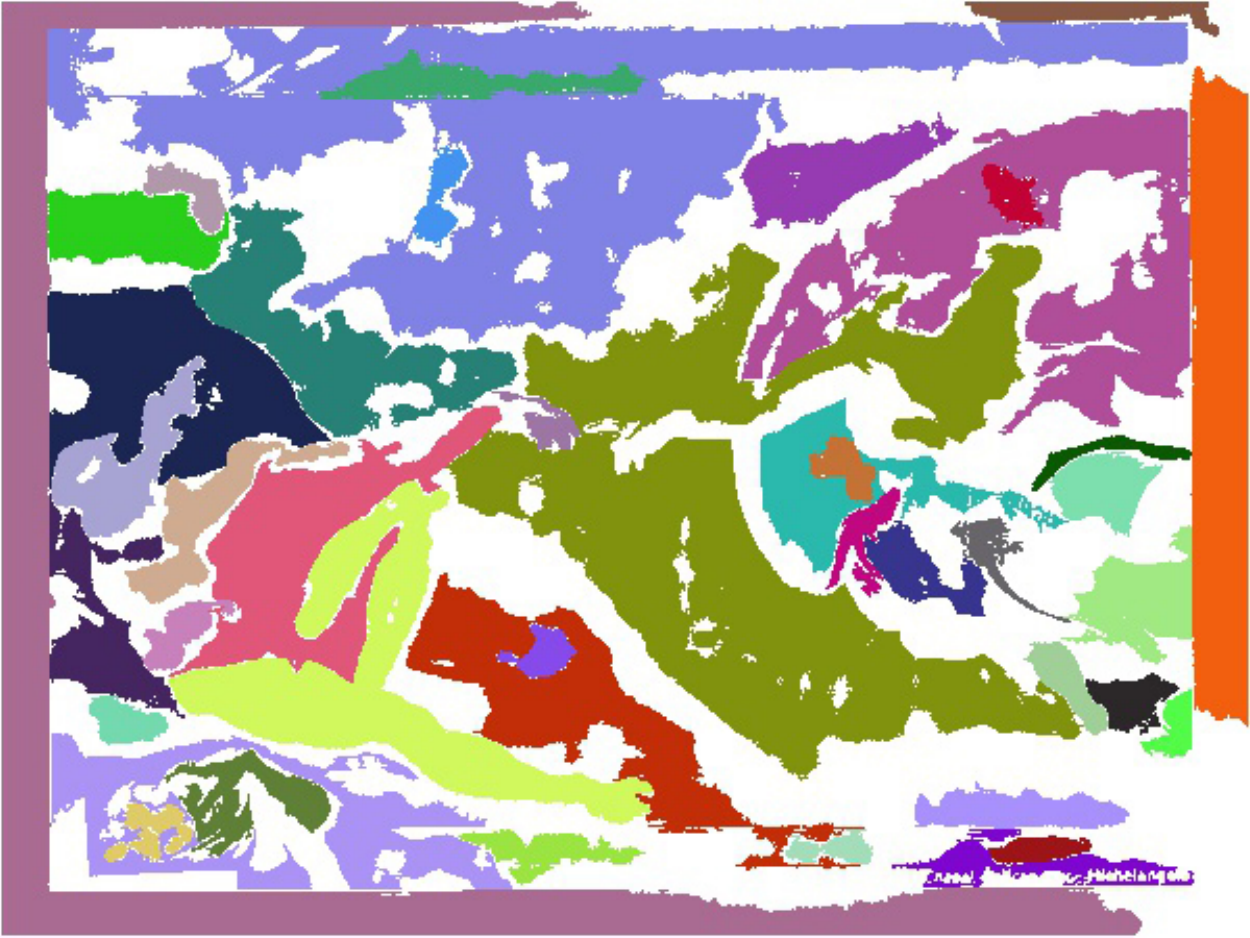}
		\caption{Region Segmentation with Image Enhancement}
		\label{Fig53}
	\end{subfigure}
	\caption{Illustration of Region Segmentation Based on Image Enhancement Effects.}
\end{figure}
\subsubsection{Region Segmentation}
The region segmentation in this paper is implemented using the Maximally Stable Extremal Regions (MSER) algorithm, an improved watershed method by Nister David \cite{ref16}. The watershed method analogizes pixel grayscale values in the image to elevation, where a grayscale threshold is set as the water level. Connected regions below this threshold form the boundaries of watersheds, and as the grayscale threshold increases, several connected regions can be obtained. The MSER algorithm then extracts maximally stable extremal regions by applying constraints based on the rate of change within the same region. The key operation involves starting from a point in the image, performing a 4-neighbors search, and establishing a set of pixels related to the current point's grayscale threshold. During the search, the pixel set is manipulated based on the threshold changes of the search point, resulting in connected regions at different grayscale thresholds. Subsequently, the maximally stable extremal regions are filtered using equation (9), and overlapping regions are merged.
\begin{equation} 
	\label{eq9}
	q(i)= \frac{|Q_{i}-Q_{i-delta}|}{Q_{i-delta}}
\end{equation}	
Where $i$ is the grayscale value threshold, $Q_{i}$ is a connected region at threshold $i$, delta represents a small change in grayscale threshold, $q_{i}$  is the rate of change of region $Q_{i}$ at threshold $i$. When this rate is below the set maximum change rate, the connected region is considered a maximally stable extremal region.

The region segmentation results are illustrated in Figure 6, where the algorithm divides the region in Figure 6(a) into segmented regions in Figure 6(b). Each color represents a distinct region, and white areas indicate undetected regions identified by the algorithm as unstable. Moreover, when facing affine transformations in the image, the algorithm can still delineate corresponding regions based on grayscale characteristics. As shown in Figure 7, even when the skew angle reaches 30 degrees in an affine transformation, the regions obtained by this segmentation method still have corresponding counterparts, as evidenced by the correspondence of the green regions in Figure 7(a) and Figure 7(b).
\begin{figure}[H]	
	\centering
	\begin{subfigure}{0.45\linewidth}
		\centering
		\includegraphics[width=0.95\linewidth]{61.pdf}
		\caption{Original Image}
		\label{Fig61}%
	\end{subfigure}
	\centering
	\begin{subfigure}{0.45\linewidth}
		\centering
		\includegraphics[width=0.95\linewidth]{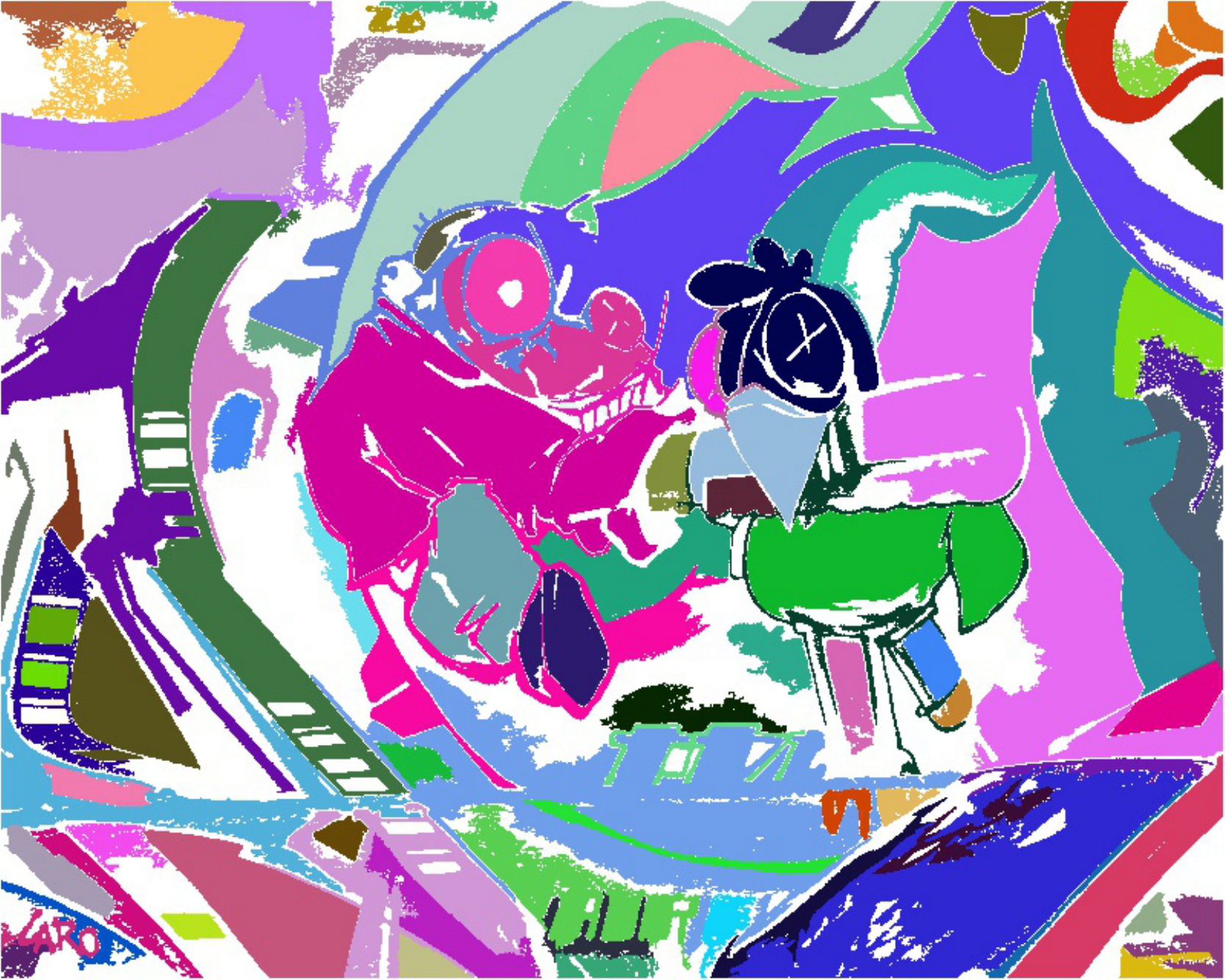}
		\caption{ MSER Region Segmentation Image}
		\label{Fig62}
	\end{subfigure}
	\caption{Illustration of Region Segmentation.}
\end{figure}

\begin{figure}[H]	
	\centering
	\begin{subfigure}{0.45\linewidth}
		\centering
		\includegraphics[width=0.95\linewidth]{61.pdf}
		\caption{Original Image}
		\label{Fig71}%
	\end{subfigure}
	\centering
	\begin{subfigure}{0.45\linewidth}
		\centering
		\includegraphics[width=0.95\linewidth]{72.pdf}
		\caption{Image with 30-degree Tilt}
		\label{Fig72}
	\end{subfigure}
	\caption{Corresponding Regions under Affine Transformation.}
\end{figure}
\subsubsection{Region Description}
The augmentation of region information in this paper is based on invariants derived from affine transformations in the image. The invariants reflected from the decomposition of the affine matrix, as in Equation (2), are as follows:

Invariant 1: Parallel lines in the image intersect at an infinite point $(x_{1},y_{2},0)^{T}$. Bringing this infinite point into the affine transformation equation (1) results in points that are also at an infinite point. Hence, lines remain parallel after the transformation.

Invariant 2: The ratio of lengths of parallel line segments remains unchanged. From Equation (2), it is evident that the length transformation of line segments is only related to the image scaling factor $\lambda$ , the affine quantity $t$, and the angle $\phi$  of the line segment relative to the positive x-axis. The scale factor is $ \lambda \sqrt{tcos^{2}\phi+sin^{2}\phi} $, ensuring the invariant proportionality of parallel line segments.

Invariant 3: The area ratio of various regions in the image remains invariant. Rotation and translation do not alter the area, and during an affine transformation, the area scaling is $\lambda ^{2}t$ times.

In addition to the three invariants mentioned above, considering only the effects of camera pose changes due to affine transformations and assuming no changes in pixel grayscale values in the image, the grayscale histogram's distribution probability is used when describing the segmented regions. This approach ensures the invariance of region description information under affine transformations. Moreover, given the uneven illumination in real-world scenes, when there is a change in camera pose, the pixel grayscale at the same position in the image may experience slight variations. To account for this, the histogram is computed using groups of every five grayscale values. This operation not only saves storage space for descriptors but also facilitates faster descriptor matching.

By incorporating the grayscale information of feature points within their respective larger regions, the issue of similar descriptors in different large regions can be effectively alleviated. However, this may lead to a reduction in the distinctiveness of descriptors for feature points within the same large region. To mitigate this, the relative positions of feature points to the centroid of their large regions are added to maintain the distinctiveness of descriptors within the same large region.

Calculating the relative positions of feature points to the centroid involves considering the positional offset caused by affine transformations. Therefore, the regions are first subjected to coordinate normalization. Using Invariant 2, which ensures the length ratio of parallel line segments remains constant, treating the region as an integral of several parallel line segments under coordinate normalization greatly mitigates the impact of scale and affine quantity on coordinate values due to affine transformations. Coordinate normalization is expressed as Equation (10).
\begin{align} 
	\label{eq10}
&norm\_x= \frac{x-min\_x}{max\_x-min\_x} \notag \\
&norm\_y= \frac{y-min\_y}{max\_y-min\_y} 
\end{align}	
Where $(x,y)$ represents the input pixel coordinates, and $(nor\_x, \\
norm\_y)$ represents the normalized coordinates. $min\_x$,$min\_y$,\\$max\_x$ and $max\_y$ are the minimum and maximum x, y coordinates of all pixels in the region.

In the case of an affine transformation affecting the same region, where pixel grayscale changes are minimal, and the ratio of line segments remains constant, the grayscale centroid of the region corresponds to the same pixel. The coordinates of the region's centroid (C) and the main orientation angle ($\theta$) of the region are determined using Equation (11).
\begin{align} 
	\label{eq11}
	&m_{pq}=\sum\begin{matrix}
		\\ x,y\epsilon S
		\end{matrix}x^{p}y^{q}I(x,y) \quad p,q={0,1} \notag \\
		&C=(\frac{m_{10}}{m_{00}},\frac{m_{01}}{m_{00}}) \notag \\
		&\theta = arctan(\frac{m_{01}}{m_{10}})
\end{align}	
Where $m_{pq}$ represents the moments of region $S$, $m_{00}$ denotes the total pixel grayscale sum of region $S$, $m_{10}$ is the grayscale moment in the x-direction for region $S$, and $m_{01}$ is the grayscale moment in the y-direction for region $S$. By reorienting the x-axis to the main orientation, the impact of rotation in the affine transformation on coordinate values is eliminated.

The relative positions of feature points obtained through the aforementioned coordinate normalization and grayscale centroid method provide a significant distinction between similar descriptors within the same region. Simultaneously, it ensures the invariance of the numerical values of feature point relative positions under affine transformations. Subsequently, the histogram distribution probabilities of feature points corresponding to the region and the relative positions of feature points to the centroid are weighted with $\alpha _{1}$ and $\alpha _{2}$, respectively. These weighted values are then incorporated into the original descriptors, resulting in a new region information descriptor proposed in this paper.

\section{Experimental and Results Analysis}
\subsection{Evaluation Metrics}
There are typically two types of evaluation metrics for feature point descriptors: one involves homography matrix evaluation, which verifies the corresponding accuracy of feature matching on two-dimensional images based on the homography matrix obtained from these descriptors; the other involves fundamental matrix evaluation, which validates the accuracy of feature matching in three-dimensional space geometric constraints based on the fundamental matrix obtained from these descriptors. This experiment will adopt both evaluation metrics to test the descriptors, providing strategic support for multi-view three-dimensional point cloud reconstruction.
\subsubsection{Homography Matrix Evaluation}
The homography matrix represents the transformation matrix between two images and can accurately verify the correspondence accuracy of feature points obtained by descriptors. The evaluation criterion used is the accuracy proposed by Mikolajczyk \cite{ref19}. It defines that if the distance between corresponding feature points after homography matrix transformation is less than a predefined threshold, the feature point correspondence is considered correct, as shown in Equation (12).
\begin{equation} 
	\label{eq12}
	\varepsilon < dist(H_{1},a,b)
\end{equation}	
In this context, $H_{1}$ represents the homography matrix for transforming image A to image B. The evaluation involves computing the distance between the coordinates of feature point a in image A and its corresponding coordinates b in image B after transformation by $H_{1}$. If this distance is less than the threshold $\varepsilon$ , the correspondence is considered correct. In this study, the threshold is set as $0.003\sqrt{w^{2}+h^{2}}$, where $w$ and $h$ are the width and height of image B.

Building upon this, considering the idea proposed by Jin \cite{ref20} that evaluation metrics should focus on the model's performance in the downstream pipeline of multi-view three-dimensional reconstruction, this experiment not only compares feature matching accuracy but also assesses the accuracy of the homography matrix obtained after this stage. This accuracy is measured by the precision of retained feature correspondences after homography matrix transformation.
\subsubsection{Fundamental Matrix Evaluation}
The fundamental matrix represents the geometric constraints satisfied by corresponding points in a three-dimensional scene between two images. Although it cannot precisely reflect the corresponding transformations of points, it is a crucial constraint for recovering camera poses in three-dimensional reconstruction. Therefore, the accuracy of descriptors in spatial geometric constraints is verified through the fundamental matrix. Corresponding points are considered correct when the symmetric epipolar distance d is less than a threshold. The accuracy is then calculated based on this criterion, and the calculation of d is given by Equation (13).
\begin{equation} 
	\label{eq13}
	d=(x^{T}Fx)^{2}\begin{bmatrix}
		\frac{1}{(Fx)_{2}^{1}+(Fx)_{2}^{2}}+\frac{1}{(F^{T}x^{'})_{2}^{1}+(F^{T}x^{'})_{2}^{2}}
		\end{bmatrix}
\end{equation}	
Where $d$ is the symmetric epipolar point distance, $F$ is the fundamental matrix from Image A to Image B, $x$ and $x'$ are the homogeneous coordinates of corresponding feature points, and $(Fx)_{2}^{1}$ represents the square of the first element of the $4\times1$ matrix obtained by multiplying $F$ and $x$. If the subscript is 2, it denotes the second element in the matrix. In this paper, a threshold of $5\times10^{-4}$ is set.
\subsection{Dataset Selection}
This paper employs two datasets, each corresponding to the two types of evaluation criteria mentioned in Section 4.1, namely the homography matrix dataset and the fundamental matrix dataset.

The homography matrix dataset primarily utilizes the Graffiti and Wall image sets from the Oxford dataset \cite{ref21}, which involve changes in viewpoints. Additionally, 10 image sets from the Hpatch dataset \cite{ref22} are selected to supplement the former, validating the effectiveness and higher accuracy of our descriptor in feature matching on images with affine transformations in various scenes. These 12 image sets each contain 6 images captured from different viewpoints. The first two image sets mainly vary the viewpoint by gradually increasing the tilt angle $\theta$ from 10 degrees to 60 degrees, increasing the level of affine transformation from low to high. However, they also include smaller rotations around the optical axis and variations where the camera is not tilted around a single point. The following 10 sets have random variations in viewpoint for each image. Additionally, this paper simulates different levels of affine transformations on three images, generating three sets of simulated image datasets. Each set simulates the affine variable $t$ on the original image, where $t\epsilon{\sqrt{2}, 2,2\sqrt{2},4,4\sqrt{2}}$. These simulated image sets provide a more intuitive reflection of the effectiveness of the descriptor under various degrees of affine transformations.

The fundamental matrix dataset mainly adopts the Temple and Dino datasets \cite{ref23} collected by Steven and others. These datasets consist of images captured from various angles around objects. Image sets capturing a complete circle around the object are selected for feature point description, and the effectiveness of the descriptor is tested with a certain image interval. By using a larger image interval, the degree of affine transformation between images can be increased. The true values provided in these fundamental matrix datasets represent the camera poses and require the calculation of the corresponding fundamental matrix F using Equation (14).
\begin{equation} 
	\label{eq14}
	F=K^{-T}[t] _{\times}RK^{T}
\end{equation}	
Where $K$ represents the camera intrinsic parameters, $t$ is the camera translation matrix, and $R$ is the camera rotation matrix.

\subsection{Experimental Parameter Analysis}
In the simulation of affine transformations for classification, it is necessary to compare the degree of affine distortion between two images. This paper adopts the simulation of the skew angle $\theta$  for the input images and cross-matches the simulated images with the original images. The number of matches is then used to judge the relative degree of affine distortion between the two images. By simulating images with different skew angles $\theta$ in the affine dataset and analyzing the matching results obtained through the SIFT algorithm and KNN matching, the experiment aims to determine the optimal value of $\theta$ that significantly indicates the relative degree of affine distortion.

Firstly, the experiment investigates the impact of simulating different skew angles $\theta$ on the number of matches in cross-matching, as shown in Figure 8(a). The line chart illustrates the results of cross-matching after simulating affine transformations with different $\theta$ angles for two input images. The "match1" line (reference group) represents the results of feature matching without any simulation for both images. "Match2" represents the results obtained by simulating only the low-affine-distortion image with $\theta$ angle and matching it with the high-affine-distortion image. "Match3" represents the results obtained by simulating only the high-affine-distortion image with the $\theta$ angle and matching it with the low-affine-distortion image. Figure 8(a) clearly demonstrates that the number of matches when simulating affine transformations on low-affine-distortion images and then cross-matching with high-affine-distortion images is consistently greater than simulating on high-affine-distortion images and cross-matching with low-affine-distortion images. Additionally, when simulating $\theta$ in the range of 15 degrees to 45 degrees, there is a noticeable difference in the number of matches between "match2" and "match3," making it useful for judging the relative degree of affine distortion in images. However, it is still necessary to explore which values of $\theta$ should be adopted when facing images with various degrees of affine distortion to ensure a clear distinction between "match2" and "match3."

Secondly, the experiment investigates the impact of simulating different skew angles for images with different degrees of affine distortion on the difference in the number of matches obtained through cross-matching. This is done to determine an appropriate value for $\theta$. The experiment results, as shown in Figure 8(b), involve conducting experiments on images with different degrees of affine distortion in the Graffiti image set and calculating the difference in matches between "match2" and "match3." The legend indicates the magnitude of the affine distortion in the matching images, with "1" having the smallest affine distortion and "5" having the largest. From the experimental results in Figure 8(b), it can be observed that for the lines corresponding to affine distortions 1 to 3, the difference in matches between "match2" and "match3" generally starts to decrease when the parameter $\theta$ is greater than 45 degrees. For lines corresponding to affine distortions 4 to 5, due to their larger affine distortion, the difference in matches fluctuates less noticeably when simulating $\theta$ in the range of 15 degrees to 75 degrees. Therefore, considering the overall performance, it is appropriate to choose $\theta$ as 45 degrees. At this value, the difference in matches between "match2" and "match3" is more pronounced, providing a better indication of the relative degree of affine distortion between two images through cross-matching simulations of skew angles.
\begin{figure}[H]	
	\centering
	\begin{subfigure}{0.48\linewidth}
		\centering
		\includegraphics[width=0.99\linewidth]{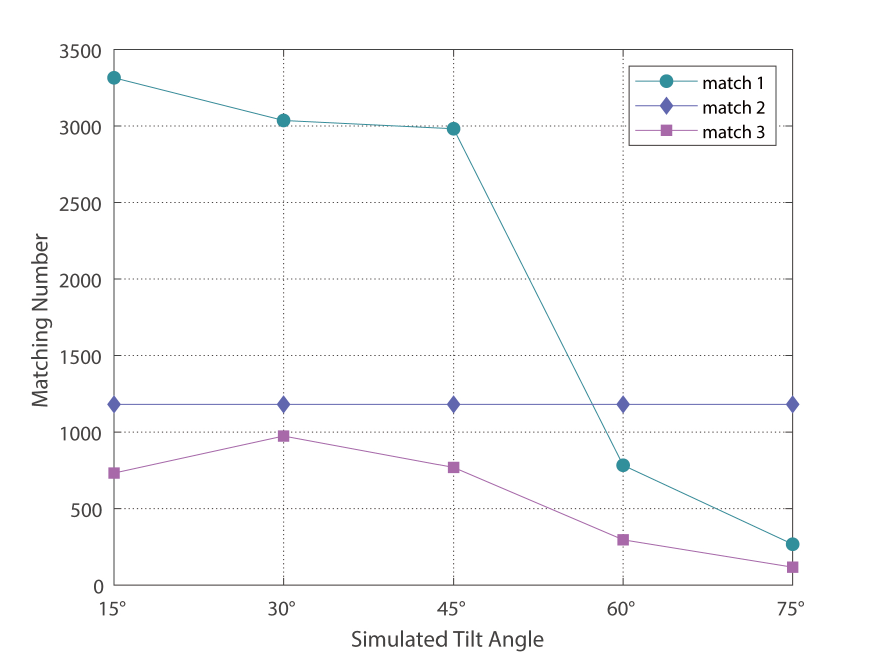}
		\caption{Number of Matches}
		\label{Fig81}%
	\end{subfigure}
	\centering
	\begin{subfigure}{0.48\linewidth}
		\centering
		\includegraphics[width=0.99\linewidth]{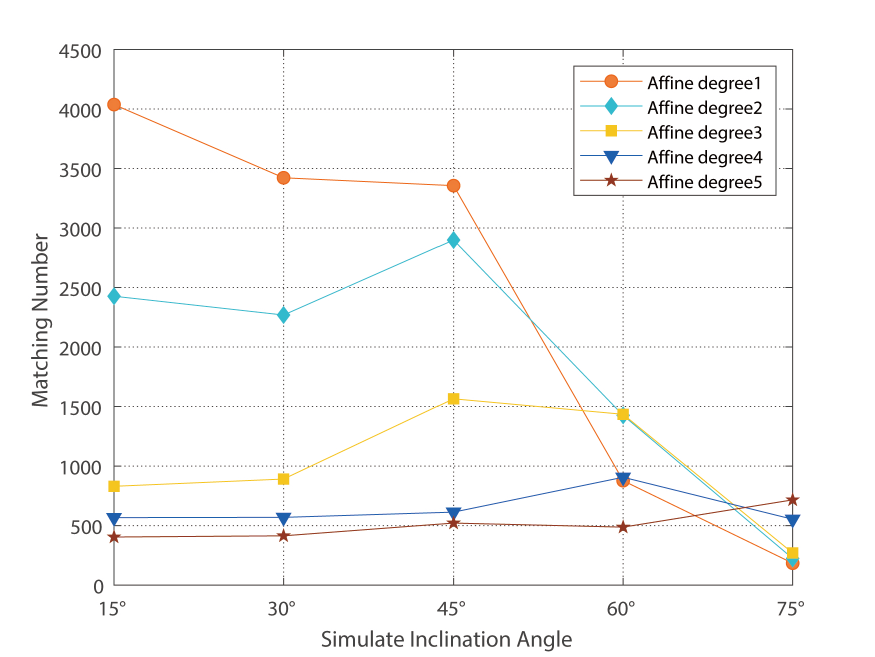}
		\caption{Difference in Number of Matches}
		\label{Fig82}
	\end{subfigure}
	\caption{Experiment on Assessing Affine Degree.}
\end{figure}

The weighting parameters $\alpha_{1}$ and $\alpha_{2}$ for the region information and the relative position of feature points have a significant impact on the algorithm's performance. Additionally, when combining this descriptor with different descriptors, the values for $\alpha_{1}$ and $\alpha_{2}$ need to be adjusted accordingly. In this parameter experiment, the original image and the image with a skew angle of 20° from the Graffiti image set were used as experimental objects. The experiments were conducted with commonly used descriptors such as SIFT, SURF, ORB, AKAZE, and BRISK.

When combining this descriptor with the SIFT descriptor, considering that the individual values of the SIFT descriptor fall within the range of (0, 256), and most values are less than 10, the magnitude of $\alpha_{1}$ and $\alpha_{2}$ in this descriptor should be set to 100 or above to effectively influence the SIFT descriptor. The experimental range was set to (100, 1000) with a step size of 100. The scatter plot in Figure 9(a) illustrates the experimental results, where the x-axis represents the parameter $\alpha_{1}$, the y-axis represents the parameter $\alpha_{2}$, and the color of the points corresponds to the accuracy of feature matching, as indicated by the color bar. The trend of color change in the scatter plot suggests that a higher accuracy is achieved when $\alpha_{2}$ is set to 300. While $\alpha_{1}$ steadily increases with the parameter, the number of matches decreases rapidly. It is recommended to set $\alpha_{1}$ to 600, as further increases have limited impact on accuracy and may sacrifice too many matches.

When combining this descriptor with the SURF descriptor, considering that the individual values of the SURF descriptor fall within the range of (0, 1), and most values are less than 0.01, the magnitude of $\alpha_{1}$ and $\alpha_{2}$ in this descriptor should be set to 0.1 or above. The experimental range was set to (0.1, 1.0) with a step size of 0.1. Based on experimental data, it is suggested to set  $\alpha_{1}$ to 0.3 and $\alpha_{2}$ to 0.1.

When combining this descriptor with the ORB descriptor, which is represented by a 256-bit binary, and the descriptor similarity is based on the Hamming distance, the magnitude of $\alpha_{1}$ and $\alpha_{2}$ should be set to 10 or above, considering that the overall values of the ORB descriptor fall within the range of 0 to 256. The experimental range was set to (10, 100) with a step size of 10. The scatter plot in Figure 9(c) presents the experimental results, and based on the data, it is recommended to set $\alpha_{1}$ to 10 and $\alpha_{2}$ to 40.

When combining this descriptor with the AKAZE descriptor, similar to the ORB descriptor, the AKAZE descriptor is represented by 488 bits binary, and the magnitude of $\alpha_{1}$ and $\alpha_{2}$ should be set to 10 or above. The experimental range was set to (10, 100) with a step size of 10. The scatter plot in Figure 9(d) illustrates the experimental results, and based on the data, it is suggested to set $\alpha_{1}$ to 10 and $\alpha_{2}$ to 60.

When combining this descriptor with the BRISK descriptor, which is represented by 512 bits binary, the magnitude of $\alpha_{1}$ and $\alpha_{2}$ should be set to 10 or above. The experimental range was set to (10, 100) with a step size of 10. The scatter plot in Figure 9(e) shows the experimental results, and based on the data, it is recommended to set  $\alpha_{1}$ to 10 and $\alpha_{2}$ to 60.

\begin{figure}[H]	
	\centering
	\begin{subfigure}{0.35\linewidth}
		\centering
		\includegraphics[width=0.95\linewidth]{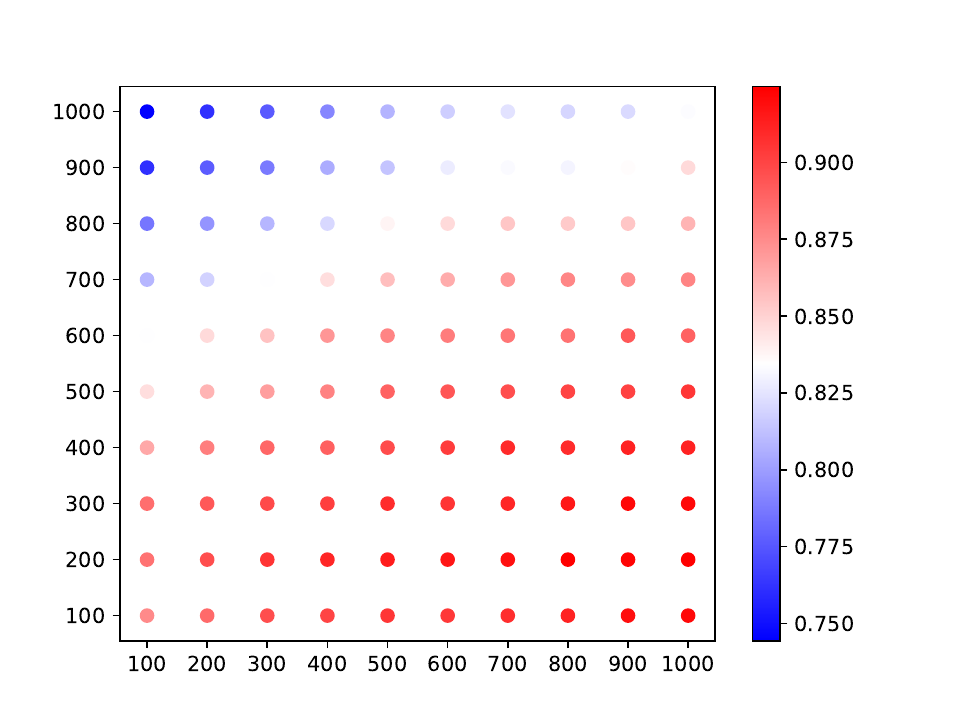}
		\caption{Combined SIFT Parameter Settings}
		\label{Fig91}%
	\end{subfigure}
	\centering
	\begin{subfigure}{0.35\linewidth}
		\centering
		\includegraphics[width=0.95\linewidth]{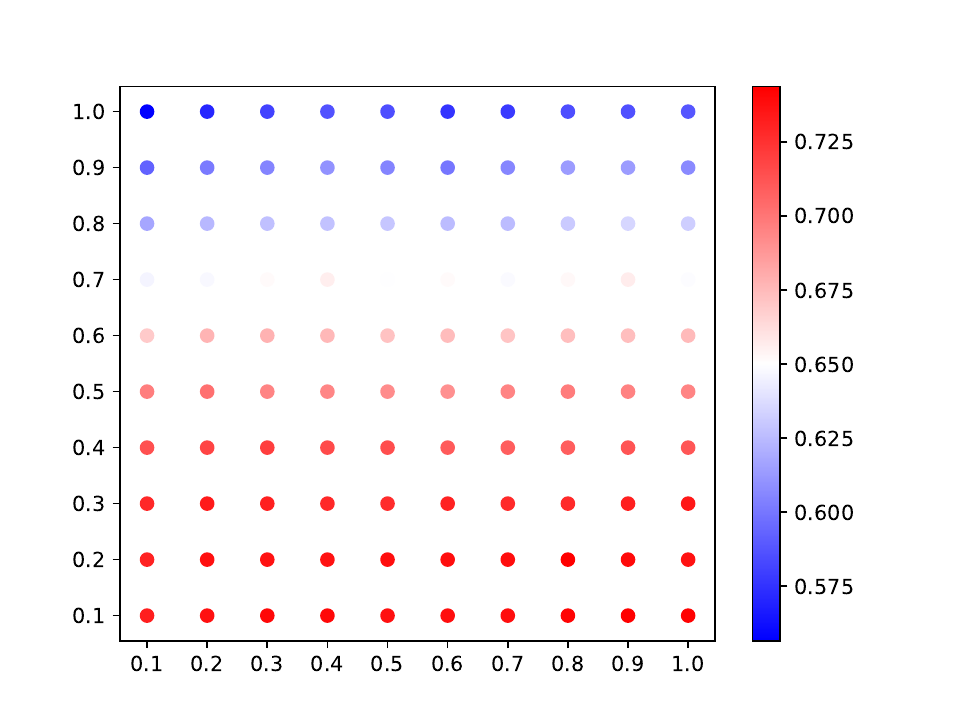}
		\caption{Combined SIFT Parameter Settings}
		\label{Fig92}
	\end{subfigure}
	\begin{subfigure}{0.35\linewidth}
		\centering
		\includegraphics[width=0.95\linewidth]{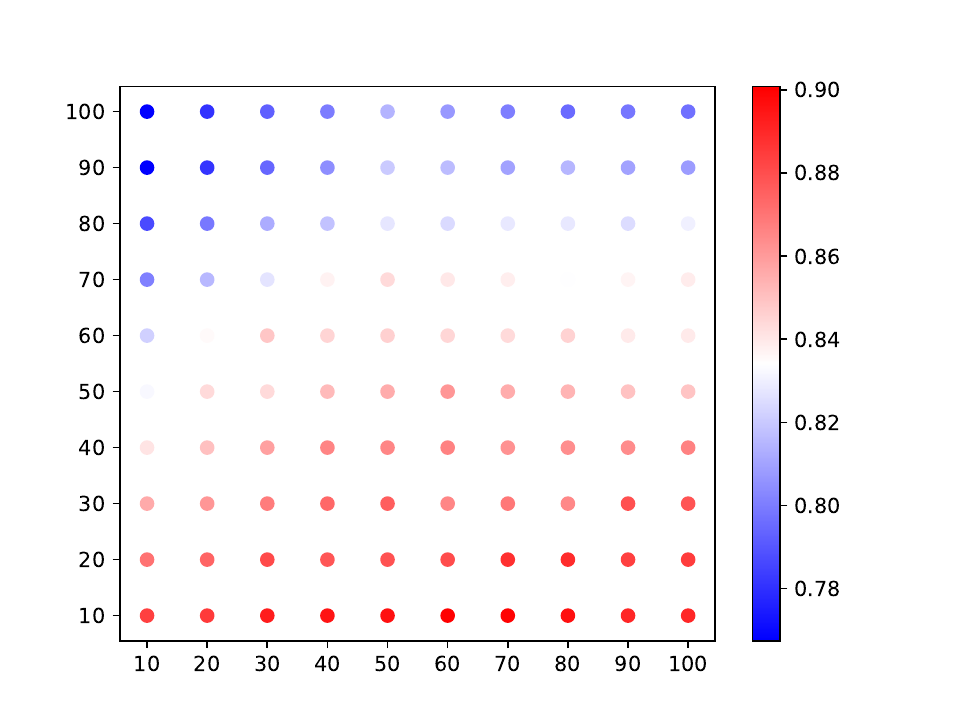}
		\caption{ Combined ORB Parameter Settings}
		\label{Fig93}
	\end{subfigure}
	\centering
	\begin{subfigure}{0.35\linewidth}
		\centering
		\includegraphics[width=0.95\linewidth]{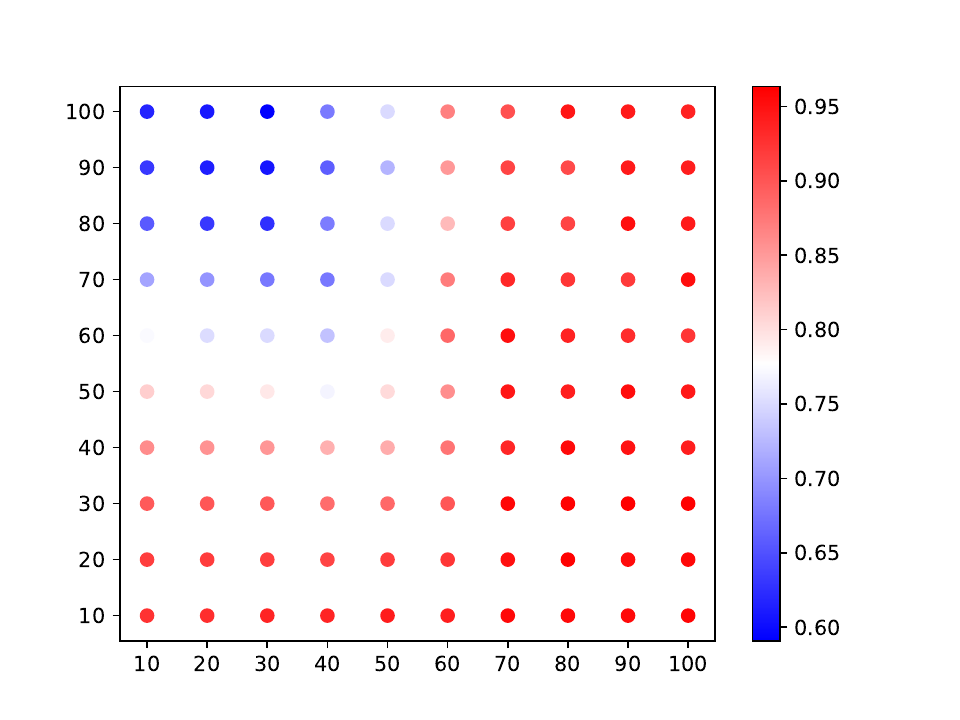}
		\caption{Combined AKAZE Parameter Settings}
		\label{Fig94}%
	\end{subfigure}
	\centering
	\begin{subfigure}{0.35\linewidth}
		\centering
		\includegraphics[width=0.95\linewidth]{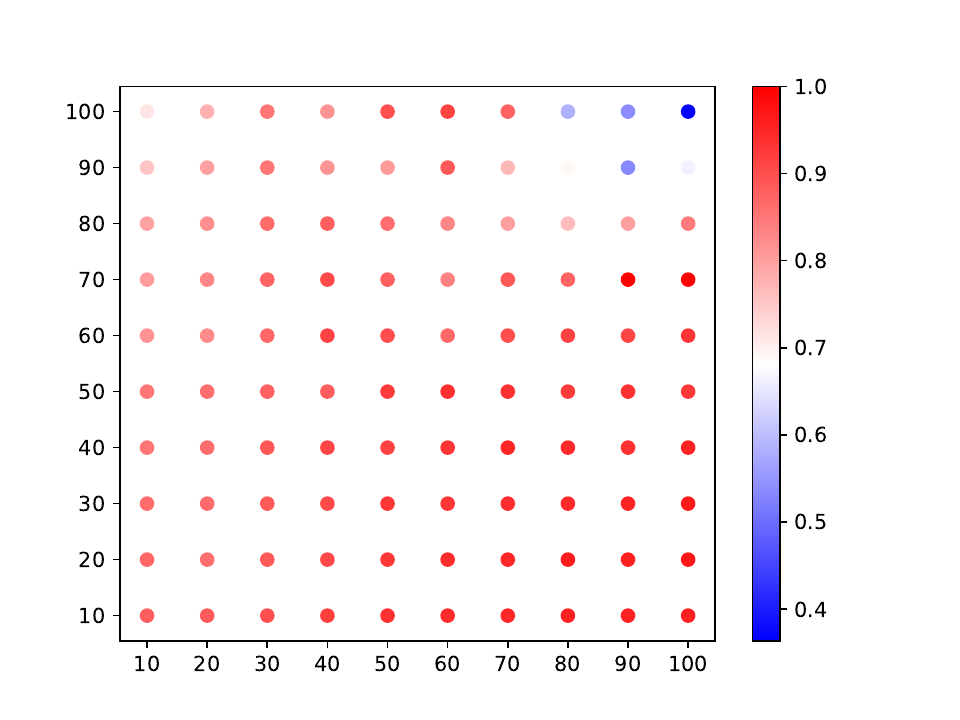}
		\caption{Combined BRISK Parameter Settings}
		\label{Fig95}
	\end{subfigure}
	\caption{ Scatter Plot of Parameters and Effects.}
\end{figure}
\subsection{Experimental Comparison}
This section presents a comparative analysis of the proposed descriptor against three classical methods: ASIFT, SIFT, and SuperPoint \cite{ref24}, focusing on the precision of feature matching. The results of the SIFT method serve as the control group, and the SuperPoint method, a deep learning approach, is included for a comparison between traditional and deep learning methods. The feature matching process involves computing the Euclidean distance between descriptors obtained by each algorithm and obtaining the corresponding relationships between the two descriptors through KNN matching. A match is considered valid only when the second distance is less than 80

(1) Comparative Experimental Evaluation Based on Homography Matrix

In experiments based on the homography matrix, evaluations were conducted on the Graffiti image set, Wall image set, and three sets of simulated image sets, as shown in Figures 10 to 11. It is observed that the proposed method exhibits higher precision and subsequent homography matrix accuracy on the Graffiti and Wall standard image sets compared to other methods, demonstrating robust precision stability. The experimental results on simulated data sets are presented in Table 1, where the proposed method outperforms others in all three simulated data sets, especially when the affine degree is large, i.e., the inclination is greater than 40 degrees. The precision of the proposed method is significantly higher than other methods. Analysis reveals that the reason for the lack of superiority over the control SIFT method in cases of small affine degree, i.e., inclination of 20 degrees, is due to the initial simulated affine quantity in the proposed descriptor and ASIFT method exceeding the original affine quantity difference between the two images. Therefore, the performance at low affine degrees is similar to the SIFT method.

For this experiment, 10 image sets were randomly selected from the Hpatch dataset for comparative analysis. The superiority of the algorithm is reflected through the average values of precision and homography matrix accuracy. The results, as shown in Table 2, indicate that the proposed method outperformed in 5 out of the 10 image sets, ASIFT method in 2 sets, SIFT method in 1 set, and SuperPoint method in 2 sets. The proposed method maintains high precision across various image sets, suggesting its adaptability to diverse scenarios.
\begin{figure}[H]	
	\centering
	\begin{subfigure}{0.48\linewidth}
		\centering
		\includegraphics[width=0.99\linewidth]{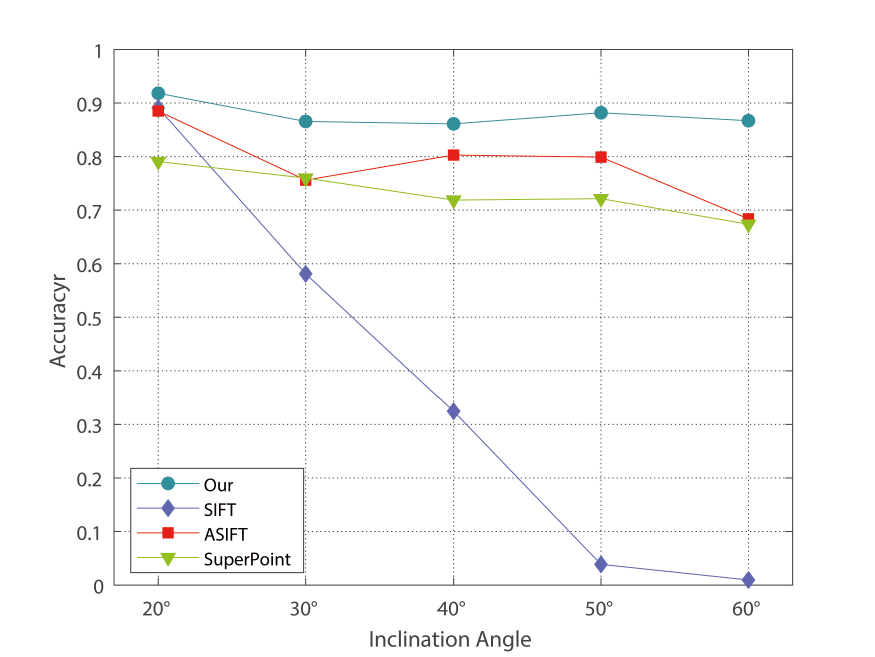}
		\caption{Feature Matching Accuracy}
		\label{Fig101}%
	\end{subfigure}
	\centering
	\begin{subfigure}{0.48\linewidth}
		\centering
		\includegraphics[width=0.99\linewidth]{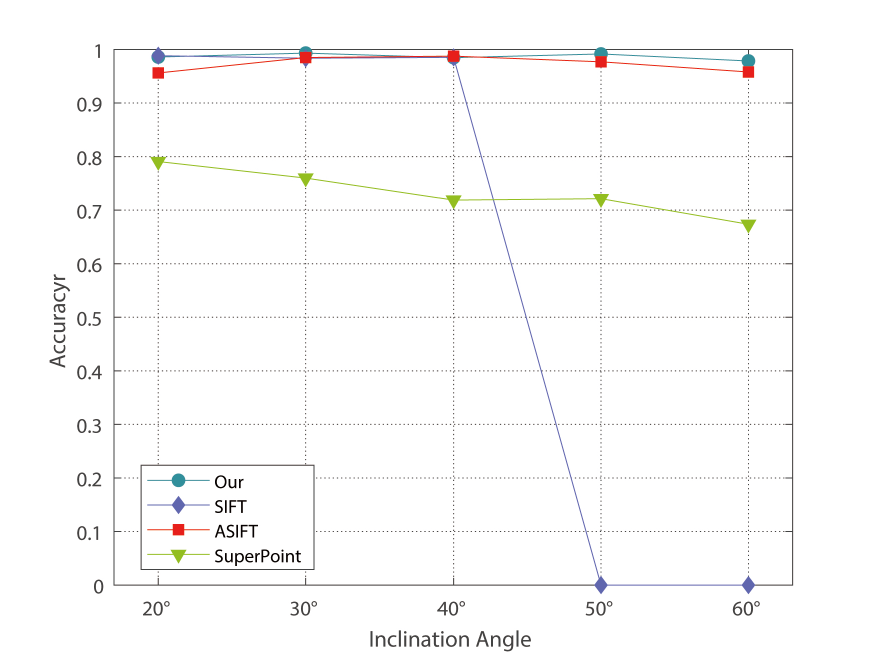}
		\caption{Homography Matrix Accuracy}
		\label{Fig102}
	\end{subfigure}
	\caption{Method Comparison on Graffiti Image Dataset.}
\end{figure}
\begin{figure}[H]	
	\centering
	\begin{subfigure}{0.48\linewidth}
		\centering
		\includegraphics[width=0.99\linewidth]{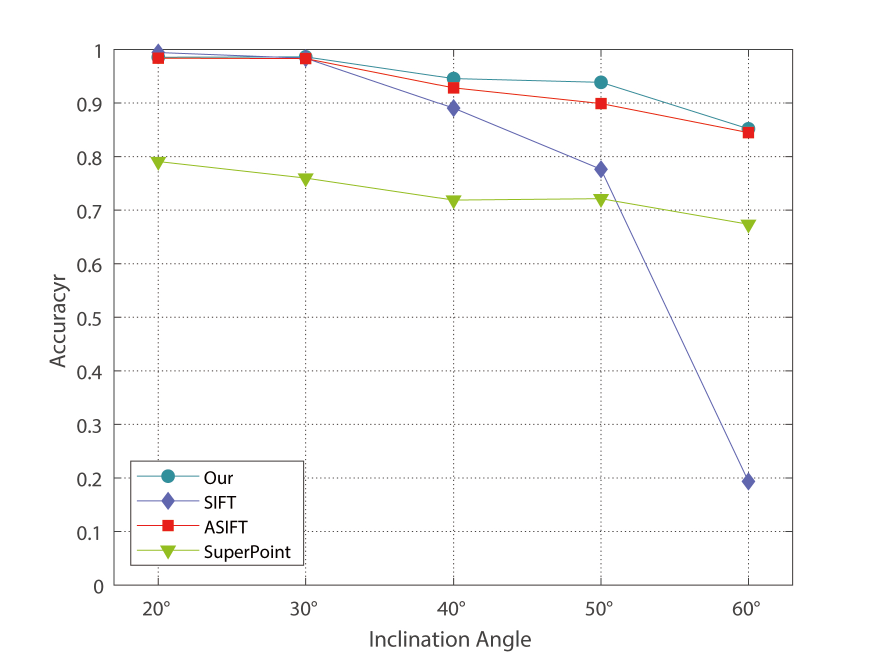}
		\caption{Feature Matching Accuracy}
		\label{Fig111}%
	\end{subfigure}
	\centering
	\begin{subfigure}{0.48\linewidth}
		\centering
		\includegraphics[width=0.99\linewidth]{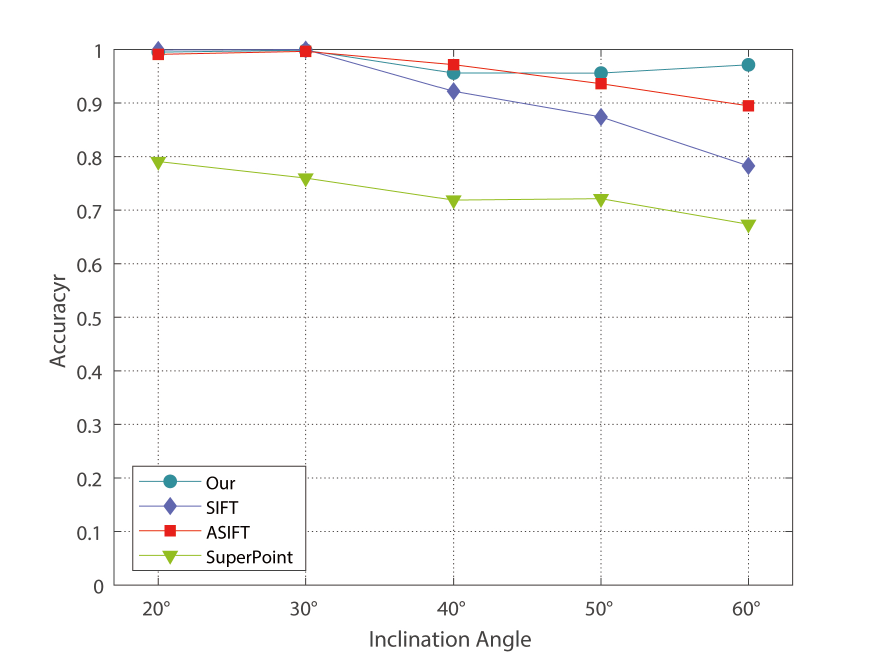}
		\caption{Homography Matrix Accuracy}
		\label{Fig112}
	\end{subfigure}
	\caption{Method Comparison on Wall Image Dataset.}
\end{figure}

\begin{table*}[ht]
	\newcommand{\tabincell}[2]{\begin{tabular}{@{}#1@{}}#2\end{tabular}}
	\centering
	\caption{Comparison of Methods on Simulated Image Sets}
	\setlength{\tabcolsep}{0.3pt}
	\label{tab1}

	\begin{tabular}{cccccccccc}
	\toprule
	\multirow{2}{*}{\tabincell{c}{Image \\ Dataset}} & \multirow{2}{*}{\tabincell{c}{Simulated Affine \\Measures}} & \multicolumn{2}{c}{Our} & \multicolumn{2}{c}{ASIFT} & \multicolumn{2}{c}{SIFT} & \multicolumn{2}{c}{SuperPoint} \\ \cmidrule(r){3-4} \cmidrule(lr){5-6} \cmidrule(lr){7-8} \cmidrule(l){9-10} 
							&                       &  Accuracy &  \tabincell{c}{Homography \\Accuracy} & Accuracy &  \tabincell{c}{Homography \\Accuracy} &Accuracy &  \tabincell{c}{Homography \\Accuracy}& Accuracy &  \tabincell{c}{Homography \\Accuracy} \\ \hline
	\multirow{5}{*}{\tabincell{c}{Simulated\\Image \\ Dataset1}}  & $\sqrt{2}$           & 90.19\%     & 99.57\% & 88.50\% & 98.24\% & 85.60\% & 99.54\% & 83.70\% & 98.37\%\\ 
							  & 2      & 92.54\%     & 98.58\% & 82.64\%  & 95.96\% & 50.82\% & 84.00\% & 66.82\% & 95.68\%\\
							  & 2$\sqrt{2}$     & 91.26 \%   & 97.84\%   &76.03\%  & 94.20\% & 15.79\% & 83.33\% & 53.57\% & 93.10\% \\ 
							  & 4 & 83.28\%     & 95.10\% & 62.42\%  & 87.28\% & 10.34\% & 0.00\% & 35.71\% & 62.50\%\\
							  & 4$\sqrt{2}$ & 78.06\%     & 95.91\% & 51.54\%  & 86.37\% & 0.00\% & 0.00\% & 16.67\% & 25.00\%\\
	\hline
	\multirow{5}{*}{\tabincell{c}{Simulated\\Image \\ Dataset2}}  & $\sqrt{2}$           & 94.74\%     & 100.00\% & 95.63\% & 99.56\% & 90.19\% & 99.70\% & 89.18\% & 98.21\%\\ 
							  & 2      & 96.49\%	&99.80\%	&93.50\%	&99.28\%	&45.65\%	&97.92\%	&91.10\%	&98.36\%\\
							  & 2$\sqrt{2}$     & 95.35\%	&99.82\%	&89.94\%	&98.36\%	&11.27\%	&81.82\%	&61.90\%	&90.48\%\\ 
							  & 4 & 91.56\%	&99.47\%	&80.75\%	&98.07\%	&2.03\% &	13.33\%	&34.78\%	&100.00\%\\
							  & 4$\sqrt{2}$ & 82.94\%	&98.97\%	&66.23\%	&92.02\%	&0.00\%	&0.00\%&	23.08\%	&25.00\%\\
	\hline
	\multirow{5}{*}{\tabincell{c}{Simulated\\Image \\ Dataset1}}  & $\sqrt{2}$           & 92.56\%	&99.15\%	&92.66\%	&98.89\%	&87.00\%	&99.32\%	&86.61\%	&99.36\%\\ 
							  & 2      & 95.13\%	&99.33\%	&88.94\%	&98.49\%	&41.88\%	&100.00\%	&85.34\%	&99.34\%\\
							  & 2$\sqrt{2}$     & 92.75\%	&98.75\%	&83.74\%	&97.89\%	&8.25\%	&0.00\%	&60.71\%	&89.29\%\\ 
							  & 4 & 85.33\%	&97.29\%	&77.18\%	&94.91\%	&2.68\%	&0.00\%	&25.00\%	&0.00\%\\
							  & 4$\sqrt{2}$ & 80.62\%	&97.90\%	&66.38\%	&91.49\%	&0.66\%	&0.00\%	&0.00\%	&0.00\%\%\\
 \bottomrule
	\end{tabular}
	\end{table*}

		\begin{table*}[ht]
			\newcommand{\tabincell}[2]{\begin{tabular}{@{}#1@{}}#2\end{tabular}}
			\centering
			\caption{Comparison of Methods on the Hpatch Image Dataset}
			\setlength{\tabcolsep}{0.5pt}
			\begin{tabular}{ccccccccc}
			\toprule
			\tabincell{c}{Image \\ Dataset} & \multicolumn{2}{c}{Our} & \multicolumn{2}{c}{ASIFT} & \multicolumn{2}{c}{SIFT} & \multicolumn{2}{c}{SuperPoint} \\
			\cmidrule(r){2-3} \cmidrule(lr){4-5} \cmidrule(lr){6-7} \cmidrule(l){8-9}
			& Accuracy &  \tabincell{c}{Homography \\Accuracy}  & Accuracy &  \tabincell{c}{Homography \\Accuracy} & Accuracy &  \tabincell{c}{Homography \\Accuracy}  &Accuracy &  \tabincell{c}{Homography \\Accuracy}  \\
			\midrule
						   v\_abstract           & 88.82\%	&99.27\%	&84.48\%	&99.12\%	&70.99\%	&95.23\%	&85.04\%	&97.93\%\\ 
										v\_adam	&87.51\%	&97.01\%	&83.01\%	&93.63\%	&74.14\%	&96.24\%	&72.73\%	&92.05\% \\
										v\_apprentices	&92.69\%	&99.77\%	&92.37\%	&99.83\%	&76.05\%	&99.71\%	&93.66\%	&99.33\%\\
										 v\_artisans	&94.36\%	&99.68\%	&92.67\%	&99.40\%	&83.88\%	&98.73\%	&88.08\%	&97.88\%\\
										  v\_astronautis	&91.90\%	&99.17\%	&91.89\%	&99.34\%	&73.77\%	&99.60\%	&92.14\%	&98.52\%\\
										 v\_azzola	&80.38\%	&92.05\%	&78.01\%	&91.07\%	&66.64\%	&91.81\%	&72.69\%	&89.85\%\\
										   v\_bark	&91.29\%	&99.63\%	&91.92\%	&99.01\%	&93.22\%	&100.00\%	&15.33\%	&18.33\%\\
										  v\_bees	&91.66\%	&98.99\%	&94.92\%	&99.18\%	&92.25\%	&99.73\%	&89.56\%	&97.43\%\\
										 v\_beyus	&89.72\%	&99.72\%	&82.38\%	&98.83\%	&51.91\%	&96.98\%	&79.18\%	&96.98\%\\
										  v\_bip	&88.87\%	&99.87\%	&92.85\%	&99.77\%	&88.24\%	&99.93\%	&56.67\%	&69.08\%\\

			\bottomrule
			\end{tabular}
			\end{table*}
(2) Comparative Experimental Evaluation Based on Fundamental Matrix

In experiments based on the fundamental matrix, the results of the comparative analysis between our method and the SIFT, ASIFT, and SuperPoint methods on the temple and Dino datasets are shown in Figure 12(a) and Figure 12(b). The experiments aimed to showcase the overall performance of the descriptor method on datasets with different affine degrees, as reflected by the average accuracy of the fundamental matrix at various image intervals. Across different image intervals, our method demonstrated a significant advantage over the traditional methods, SIFT and ASIFT, only slightly lagging behind the deep learning-based SuperPoint method. This is attributed to the fact that the loss function in the SuperPoint network is based on the fundamental matrix, which better aligns with the accuracy evaluation of feature matching in three-dimensional space. SuperPoint's approach is advantageous for three-dimensional reconstruction, requiring compliance with corresponding geometric constraints in three-dimensional space, but may not fully satisfy precise point correspondence on two-dimensional images. On the other hand, our method, being a traditional approach, focuses solely on identifying corresponding information in two-dimensional images, lacking constraints from three-dimensional space. This increases the probability of errors in the corresponding relationships within the fundamental matrix. Therefore, under this evaluation criterion, our algorithm slightly lags behind deep learning methods, but the difference is relatively small.

\begin{figure}[H]	
	\centering
	\begin{subfigure}{0.35\linewidth}
		\centering
		\includegraphics[width=0.99\linewidth]{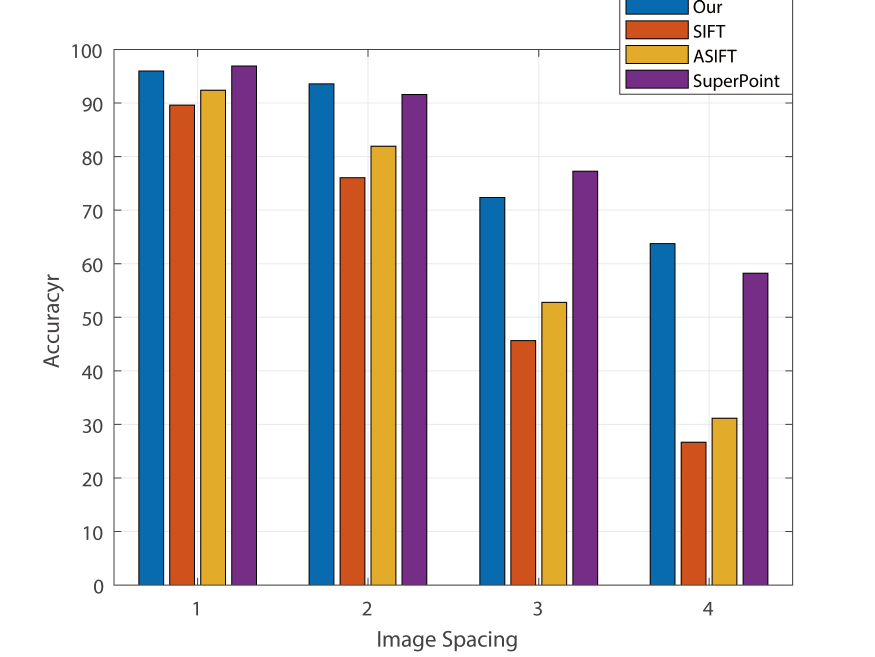}
		\caption{Method Comparison on Temple Image Dataset}
		\label{Fig121}%
	\end{subfigure}
	\centering
	\begin{subfigure}{0.35\linewidth}
		\centering
		\includegraphics[width=0.99\linewidth]{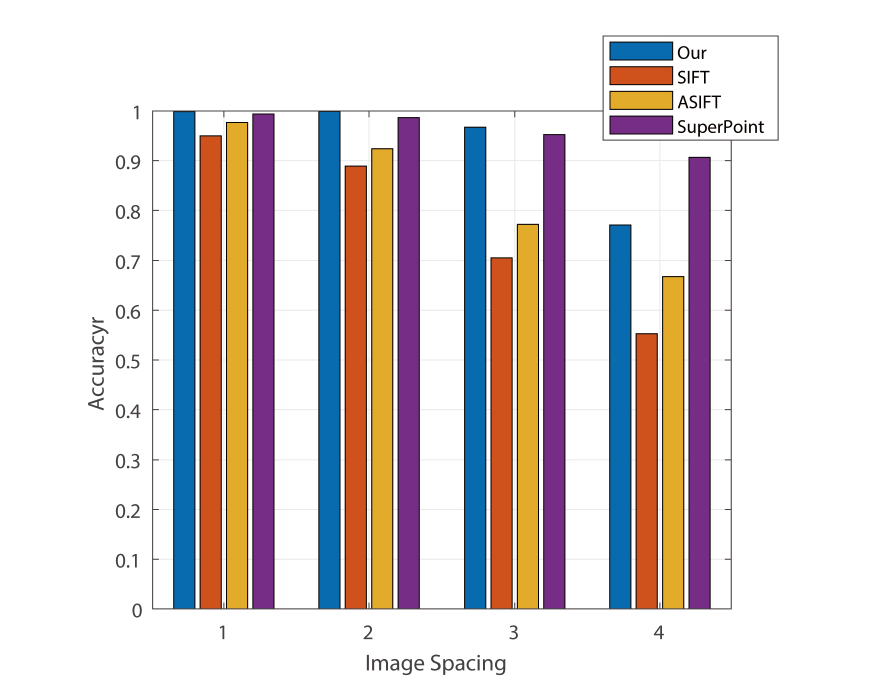}
		\caption{Method Comparison on Dino Image Dataset}
		\label{Fig122}
	\end{subfigure}
	\caption{Experiment Results on Fundamental Matrix Accuracy Comparison.}
\end{figure}
(3) Descriptor Compatibility Experiment

This section conducts experiments to test the compatibility of our method with other descriptors, specifically evaluating the matching accuracy when our method is combined with various descriptors. Five descriptors, namely SIFT, SURF, ORB, AKAZE, and BRISK, are chosen for experimentation. The Graffiti image set is utilized as the dataset, comparing the matching accuracy of the original descriptors, the combination of original descriptors with our method, and the combination of original descriptors with an Affine affine simulation algorithm under the evaluation criterion of the homography matrix. The experimental results are presented in Figure 13. Figure 13 illustrates the matching accuracy of the descriptors SIFT, SURF, ORB, AKAZE, and BRISK when combined with our method under different degrees of affine transformations. The results are compared with the original descriptors and the combination of original descriptors with the Affine affine simulation algorithm. In the graph, the vertical axis represents accuracy, and the horizontal axis represents the inclination angle, where a larger inclination angle indicates a greater degree of affine transformation. The legend "Our" denotes the combination of our method with the original descriptors, "Affine" represents the combination with the Affine affine simulation algorithm, and "Origin" represents the original descriptors. From Figure 13, it can be observed that our method maintains a consistently high and stable accuracy level across different degrees of affine transformations. Moreover, the accuracy of our method is noticeably higher than that of the original descriptors and descriptors combined with the Affine algorithm under various affine changes.
\begin{figure*}[t]
    \centering
    \includegraphics[height=1.8in]{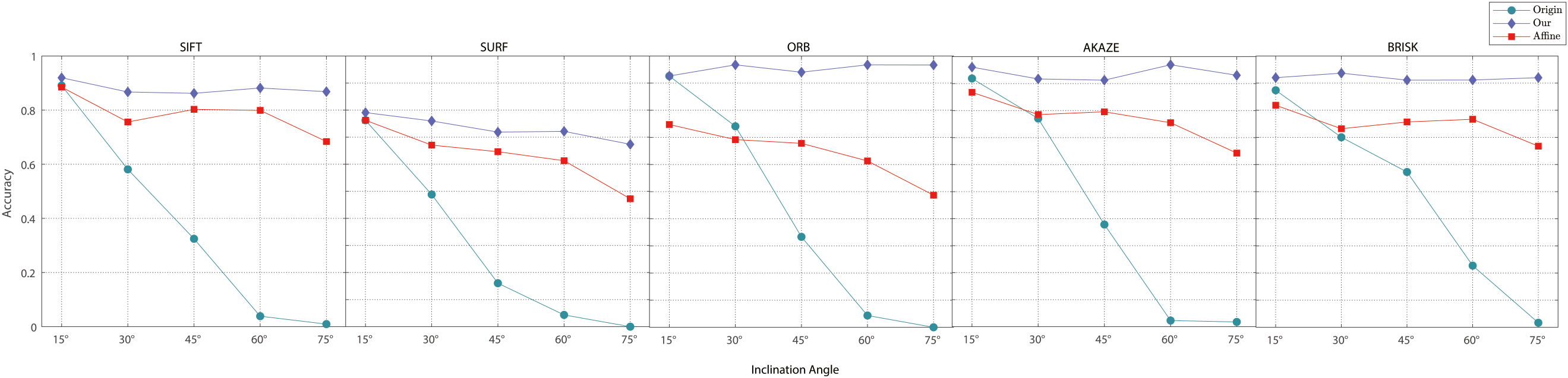}
\caption{Line chart comparing the adaptability of the proposed method to commonly used descriptors in the literature.}
    \label{Fig13}
\end{figure*}

(4) Runtime Comparison

The experimental setup for this study is as follows:

The main hardware includes an Intel(R) Core(TM) i7-10875H CPU, 16GB of RAM, and an NVIDIA GeForce RTX 2060 6GB graphics card. The software environment consists of PyCharm Professional 2020.1, with dependencies primarily on the OpenCV-Python library version 4.5.3 and the NumPy library version 1.19.2.

Under these running conditions, a comparison of the construction time and matching time of descriptors between the proposed descriptor, SIFT method, and ASIFT method is conducted. The average times for the Graffiti image set experiment are presented in Table 3.

\begin{table}[!hptb]
	\centering
	\caption{Running time}
	\setlength{\tabcolsep}{0.3pt}
	\begin{tabular} {l *{2}{S[table-format=2.4]}}  
	  \toprule
	  Method & {Constructing descriptors} & {Descriptor matching} \\
	  \midrule
	  SIFT & 0.588s & 0.895s \\
	  ASIFT & 4.484s & 37.884s\\
	  proposed & 7.691s & 30.6543s  \\
	
	  \bottomrule
	\end{tabular}
	
	\label{tab3}
  \end{table}    
The ASIFT method and the approach presented in this paper take significantly more time than SIFT. This is due to the simulation of affine transformations performed to ensure the effectiveness of descriptors in affine scenes, resulting in the calculation of descriptors on multiple images. Moreover, both ASIFT and the method described in this paper generate a much higher number of descriptors compared to SIFT feature points. In order to better suit affine transformations, the approach in this paper includes the calculation of information about the image regions corresponding to the feature points, requiring an average of approximately 5.773 seconds for image region segmentation. However, the optimization of strategies for simulating affine transformations in this paper results in a significantly lower total time for descriptor matching compared to the ASIFT method. 

\section{Conclusion}
This paper analyzes the reasons why commonly used descriptors fail to maintain invariance and distinctiveness in affine scenes. It proposes a region-based information feature descriptor that simulates affine transformations through classification, effectively addressing the issue of instability in the accuracy of existing descriptors under large radiometric transformations. The key focus of this method is to apply strategies for magnifying or reducing affine transformations on images with varying degrees of affine distortion. This augmentation, in addition to the original feature descriptors, includes information such as the grayscale histogram of the region to which a feature point belongs and the relative centroid position of the feature point. This significantly enhances the adaptability of the descriptors. Through extensive experiments, the paper thoroughly validates the superiority of the proposed descriptor under larger affine variations and its compatibility when combined with various commonly used descriptors.





\end{document}